\documentclass[11pt]{article}
\usepackage{chicagob}
\usepackage{latexsym}
\usepackage{eepic}

\newenvironment{oldthm}[1]{\par\noindent{\bf Theorem #1:} \em \noindent}{\par}
\newenvironment{oldlem}[1]{\par\noindent{\bf Lemma #1:} \em \noindent}{\par}
\newenvironment{oldcor}[1]{\par\noindent{\bf Corollary #1:} \em \noindent}{\par}
\newenvironment{oldpro}[1]{\par\noindent{\bf Proposition #1:} \em \noindent}{\par}
\newcommand{\othm}[1]{\begin{oldthm}{\ref{#1}}}
\newcommand{\eothm}{\end{oldthm} \medskip}
\newcommand{\olem}[1]{\begin{oldlem}{\ref{#1}}}
\newcommand{\eolem}{\end{oldlem} \medskip}
\newcommand{\ocor}[1]{\begin{oldcor}{\ref{#1}}}
\newcommand{\eocor}{\end{oldcor} \medskip}
\newcommand{\opro}[1]{\begin{oldpro}{\ref{#1}}}
\newcommand{\eopro}{\end{oldpro} \medskip}

\usepackage{amssymb}

\newcommand{\uaicommentout}[1]{}
\newcommand{\commentout}[1]{}
\newcommand{\ijcai}[1]{}
\newcommand{\BB}[1]{#1}

\newcommand{\sedom}{\mbb{E}}

\newcommand{\prfcommentout}[1]{}
\newtheorem{THEOREM}{Theorem}[section]
\newenvironment{theorem}{\begin{THEOREM}}%
                        {\end{THEOREM}}
\newtheorem{LEMMA}[THEOREM]{Lemma}
\newenvironment{lemma}{\begin{LEMMA}}%
                      {\end{LEMMA}}
\newtheorem{COROLLARY}[THEOREM]{Corollary}
\newenvironment{corollary}{\begin{COROLLARY}}%
                          {\end{COROLLARY}}
\newtheorem{PROPOSITION}[THEOREM]{Proposition}
\newenvironment{proposition}{\begin{PROPOSITION}}%
                            {\end{PROPOSITION}}
\newtheorem{DEFINITION}[THEOREM]{Definition}
\newenvironment{definition}{\begin{DEFINITION}\rm}%
                            {\end{DEFINITION}}
\newtheorem{CLAIM}[THEOREM]{Claim}
\newenvironment{claim}{\begin{CLAIM}\rm}%
                            {\end{CLAIM}}
\newtheorem{EXAMPLE}[THEOREM]{Example}
\newenvironment{example}{\begin{EXAMPLE}\rm}%
                            {\end{EXAMPLE}}
\newtheorem{REMARK}[THEOREM]{Remark}
\newenvironment{remark}{\begin{REMARK}\rm}%
                            {\end{REMARK}}

                      {}

\newcommand{\thm}{\begin{theorem}}
\newcommand{\lem}{\begin{lemma}}
\newcommand{\pro}{\begin{proposition}}
\newcommand{\dfn}{\begin{definition}}
\newcommand{\rem}{\begin{remark}}
\newcommand{\xam}{\begin{example}}
\newcommand{\cor}{\begin{corollary}}
\newcommand{\prf}{\trivlist \item[\hskip \labelsep{\bf Proof:}]}
\newcommand{\ethm}{\end{theorem}}
\newcommand{\elem}{\end{lemma}}
\newcommand{\epro}{\end{proposition}}
\newcommand{\edfn}{\bbox\end{definition}}
\newcommand{\erem}{\bbox\end{remark}}
\newcommand{\exam}{\bbox\end{example}}
\newcommand{\ecor}{\end{corollary}}
\newcommand{\eprf}{\bbox\endtrivlist}
\newcommand{\beqn}{\begin{equation}}
\newcommand{\eeqn}{\end{equation}}

\newcommand{\beqna}{\begin{eqnarray}}
\newcommand{\eeqna}{\end{eqnarray}}

\newcommand{\bbox}{\vrule height7pt width4pt depth1pt}

\newcommand{\clm}{\begin{claim}}
\newcommand{\eclm}{\end{claim}}
\newcommand{\inter}{\cap}
\newcommand{\IR}{\mbb{R}}

\renewcommand{\phi}{\varphi}
\newcommand{\A}{{\cal A}}

\newcommand{\E}{{\cal E}}

\renewcommand{\P}{{\cal P}}

\newcommand{\st}{\  \vert \ } %

\newcommand{\eg}{e.g.,}
\newcommand{\ie}{i.e.,}

\newcommand{\ol}{\setlength{\itemsep}{0pt}\begin{enumerate}}
\newcommand{\eol}{\end{enumerate}\setlength{\itemsep}{-\parsep}}
\newcommand{\ul}{\setlength{\itemsep}{0pt}\begin{itemize}}
\newcommand{\dl}{\setlength{\itemsep}{0pt}\begin{description}}
\newcommand{\edl}{\end{description}\setlength{\itemsep}{-\parsep}}
\newcommand{\eul}{\end{itemize}\setlength{\itemsep}{-\parsep}}

\newcommand{\bi}{\begin{itemize}}
\newcommand{\ei}{\end{itemize}}
\newcommand{\be}{\begin{enumerate}}
\newcommand{\ee}{\end{enumerate}}

\newcommand{\sbset}{\subseteq}

\newcommand{\ev}[1]{[\![{#1}]\!]}

\newcommand{\mfig}[1]{{Figure~\ref{#1}}}

\newcommand{\mthm}[1]{{Theorem~\ref{#1}}}
\newcommand{\mcor}[1]{{Corollary~\ref{#1}}}

\newcommand{\meqn}[1]{{(\ref{#1})}}

\newcommand{\blst}{\begin{list}{}{}}
\newcommand{\elst}{\end{list}}
\newcommand{\ben}{\begin{enumerate}}
\newcommand{\een}{\end{enumerate}}
\newcommand{\bit}{\begin{itemize}}
\newcommand{\eit}{\end{itemize}}
\newcommand{\bl}{\item}

\newcommand{\mc}[1]{\mathcal{#1}}
\newcommand{\mr}[1]{\mathrm{#1}}
\newcommand{\mi}[1]{\mathit{#1}}
\newcommand{\mb}[1]{\mathbf{#1}}

\newcommand{\mbb}[1]{\mathbb{#1}}

\newcommand{\ttt}[1]{\texttt{#1}}

\newcommand{\bfig}{\begin{figure}}
\newcommand{\efig}{\end{figure}}

\newcommand{\rar}{\rightarrow}

\newcommand{\bydef}[1]{#1}

\newcommand{\ds}{\displaystyle}

\newcommand{\eset}{\emptyset}

\newcommand{\act}{a}
\newcommand{\actb}{b}
\newcommand{\stt}{s}

\newcommand{\csq}{c}
\newcommand{\csqd}{d}

\newcommand{\Act}{A}
\newcommand{\Stt}{S}

\newcommand{\Csq}{C}

\newcommand{\ACT}{\mc{A}}

\renewcommand{\E}{\mb{E}}

\newcommand{\urv}{utility random variable}

\def\bbbu{{\mathchoice {\setbox0=\hbox{$\displaystyle\rm U$}\hbox{\hbox
to0pt{\kern0.4\wd0\vrule height1\ht0\hss}\box0}}
{\setbox0=\hbox{$\textstyle\rm U$}\hbox{\hbox
to0pt{\kern0.4\wd0\vrule height1\ht0\hss}\box0}}
{\setbox0=\hbox{$\scriptstyle\rm U$}\hbox{\hbox
to0pt{\kern0.4\wd0\vrule height1\ht0\hss}\box0}}
{\setbox0=\hbox{$\scriptscriptstyle\rm U$}\hbox{\hbox
to0pt{\kern0.4\wd0\vrule height1\ht0\hss}\box0}}}}

\def\bbbo{{\mathchoice {\setbox0=\hbox{$\displaystyle\rm O$}\hbox{\hbox
to0pt{\kern0.27\wd0\vrule height0.9\ht0\hss}\box0}}
{\setbox0=\hbox{$\textstyle\rm O$}\hbox{\hbox
to0pt{\kern0.27\wd0\vrule height0.9\ht0\hss}\box0}}
{\setbox0=\hbox{$\scriptstyle\rm O$}\hbox{\hbox
to0pt{\kern0.27\wd0\vrule height0.9\ht0\hss}\box0}}
{\setbox0=\hbox{$\scriptscriptstyle\rm O$}\hbox{\hbox
to0pt{\kern0.27\wd0\vrule height0.9\ht0\hss}\box0}}}}

\newcommand{\utf}{\mb{u}}

\newcommand{\realNum}{\mbb{R}}

\newcommand{\csqf}{\mb{c}}

\newcommand{\ep}{\oplus}
\newcommand{\EP}{\Oplus}
\newcommand{\emi}{\ominus}
\newcommand{\et}{\otimes}

\newcommand{\eleq}{\precsim}

\newcommand{\eeq}{\sim}
\newcommand{\els}{\prec}

\newcommand{\dprbsrel}{\dprbs_{\mr{pref}}}

\newcommand{\rst}{\,{\upharpoonright}\,}

\newcommand{\Oplus}{\bigoplus}

\newcommand{\axn}[1]{\mbox{A}{#1}}

\newcommand{\bottom}{\bot}

\newcommand{\notincomp}{\simeq}

\newcommand{\comp}[1]{\overline{#1}}

\newcommand{\plf}{\mr{Pl}}

\newcommand{\plstic}{plausibilistic}

\newcommand{\edom}{E}
\newcommand{\geqact}{\succsim_{\Act}}
\newcommand{\leqact}{\precsim_{\Act}}
\newcommand{\lsact}{\prec_{\Act}}

\newcommand{\lsb}{<_{2}}

\newcommand{\pdom}{P}
\newcommand{\udom}{U}

\newcommand{\ran}{\mr{ran}}

\newcommand{\paper}{paper}

\newcommand{\setthmnum}[2]{%
\setcounter{section}{#1}%
\setcounter{THEOREM}{#2}%
}

\newcounter{secStack}
\newcounter{thmStack}

\newcommand{\pushsec}{\setcounter{secStack}{\value{section}}}
\newcommand{\popsec}{\setcounter{section}{\value{secStack}}}
\newcommand{\pushthm}{\setcounter{thmStack}{\value{THEOREM}}}
\newcommand{\popthm}{\setcounter{THEOREM}{\value{thmStack}}}
\newcommand{\setThm}[2]{\pushsec\pushthm\renewcommand{\thesection}{\arabic{section}}\setthmnum{\value{#1}}{\value{#2}}}
\newcommand{\unsetThm}{\popsec\popthm\renewcommand{\thesection}{\Alph{section}}}

\newcommand{\dr}{\mc{R}}
\newcommand{\dm}{\mc{P}}

\newcommand{\dprb}{\mc{D}}

\newcommand{\dprbs}{\Pi}

\newcommand{\pldom}{\pdom}
\newcommand{\eudom}{V}

\newcommand{\pleq}{\preceq}

\newcommand{\leqRPdp}[2]{\precsim_{\mc{#1}({#2})}}
\newcommand{\lsRPdp}[2]{\prec_{\mc{#1}({#2})}}
\newcommand{\eeqRPdp}[2]{\eeq_{\mc{#1}({#2})}}

\newcommand{\gleq}{\leqRPdp{\mr{\gseu}}{\dprb}}
\newcommand{\gls}{\lsRPdp{\mr{\gseu}}{\dprb}}

\newcommand{\epupar}[3]{({#1}, {#2}, {#3})}
\newcommand{\epu}{\epupar{\edom}{\plf}{\utf}}

\newcommand{\subrules}{subrules}

\newcommand{\drule}{decision rule}

\newcommand{\drules}{decision rules}

\newcommand{\expstr}{expectation domain}
\newcommand{\anexpstr}{an expectation domain}
\newcommand{\expstrs}{expectation domains}

\newcommand{\seu}{EU}
\newcommand{\gseu}{GEU}

\newcommand{\mgseu}{Maximizing \gseu}

\newcommand{\reg}{{Mini\-max Regret}}

\newcommand{\vonNM}{\mbox{von~Neumann} and \mbox{Morgenstern}}

\newcommand{\alternatives}{alternatives}

\newcommand{\npsub}{nonempty proper subset}
\newcommand{\npsubs}{nonempty proper subsets}

\newcommand{\represent}{represent}
\newcommand{\representation}{representation}
\newcommand{\representations}{representations}
\newcommand{\representable}{representable}
\newcommand{\representability}{representability}
\newcommand{\represents}{represents}
\newcommand{\representing}{representing}
\newcommand{\represented}{represented}

\newcommand{\consequence}{consequence}
\newcommand{\consequences}{consequences}

\newcommand{\states}{states}

\newcommand{\ssofn}{states of the world}

\newcommand{\prefrel}{preference relation}
\newcommand{\prefrels}{preference relations}

\newcommand{\relate}{relate}

\newcommand{\relating}{relating}

\newcommand{\satisfies}{satisfies}
\newcommand{\satisfy}{satisfy}

\newcommand{\dprobs}{decision problems}

\newcommand{\dprob}{decision problem}
\newcommand{\dsitu}{decision situation}
\newcommand{\dsitus}{decision situations}

\newcommand{\acts}{acts}

\newcommand{\tandb}{tastes and beliefs}
\newcommand{\dmakers}{DMs}
\newcommand{\dmaker}{DM}

\newcommand{\plstr}{plausibility domain}
\newcommand{\plstrs}{plausibility domains}
\newcommand{\plmsr}{plausibility measure}
\newcommand{\plmsrs}{plausibility measures}

\newcommand{\utstr}{utility domain}

\newcommand{\utstrs}{utility domains}
\newcommand{\utfn}{utility function}
\newcommand{\utfns}{utility functions}

\newcommand{\valstr}{valuation domain}

\newcommand{\preorder}{preorder}

\newcommand{\Representing}{Representing}

\newcommand{\eprut}[1]{\E_{\Pr}(\utf_{#1})}
\newcommand{\eplut}[1]{\E_{\plf,\edom}(\utf_{#1})}

\newcommand{\deprut}[1]{\mbox{[[DELETE]]}}
\newcommand{\deplut}[1]{\mbox{[[DELETE]]}}

\newcommand{\eplutcond}[2]{\E_{\plf,\edom}({\utf_{#1}} \rst {#2})}
\newcommand{\lotto}[3]{\langle {#1}, {#2}, {#3} \rangle}

\newcommand{\down}[1]{\leavevmode\raise-.25ex\hbox{#1}}
\newcommand{\ldelim}{$\!${\down{\Large \ttt{[}}}}
\newcommand{\rdelim}{{\down{\Large \ttt{]}}}$\!$}
\newcommand{\cond}[1]{\ldelim{#1}\rdelim}
\newcommand{\epmono}{monotonic}

\newcommand{\dpadd}{additive}

\newcommand{\incomplete}{incomplete}
\newcommand{\whole}{whole}
\newcommand{\Wholeness}{Wholeness}
\newcommand{\geeq}{\eeqRPdp{\mr{\gseu}}{\dprb}}

\newcommand{\ulotto}[3]{\langle\!\langle {#1}, {#2}, {#3} \rangle\!\rangle}

\newcommand{\eua}{\alpha}
\newcommand{\eub}{\beta}

\newcommand{\Eut}{\mc{E}}

\newcommand{\Pref}{\P}

\begin{document}

\title{Great Expectations. 
\\
Part I\@: 
On the Customizability of Generalized Expected Utility%
\thanks{Work supported in part by NSF under grants IIS-0090145 and 
CTC-0208535 and by the DoD Multidisciplinary University Research
Initiative (MURI) program administered by ONR under
grant N00014-01-1-0795.}
\author{Francis C. Chu 
\and
\commentout{
Joseph Y. Halpern\\
Department of Computer Science\\
Cornell University\\
Ithaca, NY 14853, U.S.A.\\
Email: \{fcc,halpern\}@cs.cornell.edu
}%
Joseph Y. Halpern}
\date{Department of Computer Science\\
Cornell University\\
Ithaca, NY 14853, U.S.A.\\
Email: \{fcc,halpern\}@cs.cornell.edu}
}

\maketitle

\begin{abstract}
We propose a 
generalization of expected utility that we call \emph{generalized EU}
(GEU),
where a decision maker's 
beliefs are
\represented\ by 
\plmsrs\ 
and the decision maker's tastes are
\represented\ by 
general (\ie\ not necessarily real-valued)
\utfns\@.
We show that
every agent, 
\mbox{``rational''}
or not, can
be modeled as a \gseu\ 
maximizer. We then show that
we can customize \gseu\ by selectively imposing just the constraints we
want.
In particular, we show how each of Savage's postulates corresponds to
constraints on \gseu\@.  
\end{abstract}

\section{Introduction}

Many \drules\ have been proposed in the literature.
Perhaps the best-known approach is based on 
maximizing expected utility (\seu),   
calculated either with respect to a given (objective) probability
measure, as done originally by \vonNM~\citeyear{vNM}, or with respect to
a probability measure constructed from a preference
order on alternatives that satisfies certain postulates, as done
originally by Savage~\citeyear{Savage}\@.  All these approaches
follow the same pattern: they
formalize the set of \alternatives\ among which the decision maker
(\dmaker) must 
choose (typically as \emph{acts}
or 
\emph{lotteries}\footnote{
Formally, given a set $\Stt$ of \ssofn\ and another
set $\Csq$ of \emph{\consequences}, an \emph{act} $\act$ is a
function from $\Stt$ to $\Csq$ that, intuitively, associates with each
state $\stt$ 
the \consequence\ of performing $\act$ in $\stt$.
A \emph{lottery} is a probability
distribution over \consequences;
intuitively, the distribution quantifies how likely 
it is that each \consequence\ occurs.}).
\commentout{
and assume that the \dmaker\ has some preferences on these
\alternatives.  
}%
They then give a set of 
assumptions 
(often called \emph{postulates} or \emph{axioms})
such that 
the \dmaker's preferences 
on the \alternatives\ 
satisfy these 
assumptions
iff
the preferences 
have an \seu\ \representation,
\commentout{
(A \prefrel\ is \emph{\representable} by
\seu\ iff 
there exist some real-valued \utfn\ and probability measure
such that the 
ordering on the \alternatives\ based on
expected utility agrees 
with the \prefrel.)   
}%
where an \emph{\seu\ \representation} of a \prefrel\ is basically a \utfn\
(and a probability measure when acts are involved) such that
the relation among the \alternatives\ based on expected utility agrees with the
\prefrel.  Moreover, they show that the representation is essentially
unique  
in that, given two \representations\ of the same \prefrel, the \utfns\
are positive affine transformations of one another and the probability
measures are equal.
Thus, if the 
preferences  
of a \dmaker\ satisfy the 
assumptions,
then
she is 
behaving
\emph{as if} 
\commentout{
she has quantified her beliefs via a
probability measure and her tastes via a real-valued \utfn,
}%
she has quantified her tastes via a real-valued \utfn\ 
(and her beliefs via a probability measure)
and she is 
\relating\ 
the \alternatives\ according to their 
expected utility.
The 
assumptions
are typically regarded as criteria for rational behavior, 
so these results also suggest that if a DM's beliefs are actually described
by a probability measure and her 
tastes
are described by a \utfn, then she
should 
\relate\ the
\alternatives\ according to their expected utility
(if she wishes to appear rational).

Despite the appeal of \seu\ maximization,
it is well known that people do not follow its tenets in general
\cite{res}.
As a result, a host of 
extensions
of \seu\ have been proposed that
accommodate some of the more systematic violations (see, for example, 
\cite{Gul1991,GS1989,GiangShenoy01,KT1979,Luce00,Quiggin93,Schmeidler89,TK92,Yaari1987}). 
Again, the typical approach in the decision theory literature has been
to prove representation theorems.
These representation theorems essentially view a \drule\ $\dr$ as a
function that maps tastes (and perhaps beliefs, depending on the rule)
to a preference relation on alternatives.  The theorem then says that,
given a rule $\dr$, there is a set $\A_{\dr}$ of assumptions about
preference orders such that a preference relation $\preceq$ satisfies
$\A_{\dr}$ iff there exists some tastes and beliefs for the agent
such that, given these as inputs, $\dr$ returns $\preceq$. 
\commentout{
if the preferences of
the DM satisfy the 
assumptions,
then she 
behaves as if she
has some tastes and beliefs (in
the case of \drules\ that assume some representation of beliefs)
such that she is relating the alternatives according to the tenets of
$\dr$, given those tastes and beliefs.
}%

Given this plethora of rules, it would be useful to have a general
framework in which to 
study
decision making.  The framework should
also let us understand the  
relationship between various \drules.  We provide such a framework in
this paper.  

The basic idea of our approach is to generalize the notion of expected utility
so that it applies in as general a context as possible.
To this end, we introduce
\emph{\expstrs}, which are 
structures consisting of
\begin{itemize}
\item  three (component) domains: a \emph{\plstr} $\pdom$, a
\emph{\utstr} $\udom$, and a  
\emph{\valstr} $\eudom$,
\item two binary operators $\ep: \eudom \times \eudom \rightarrow
\eudom$ and  
$\et: \pdom \times \udom \rightarrow \eudom$,
which are the analogues of $+$ and $\times$ over the 
reals, and 
\item a reflexive binary relation
$\eleq$ 
on $\eudom$
(which generalizes $\leq$).
\end{itemize}
Intuitively, $\et$ combines plausibility values and utility values much
the same way that $\times$ combines probability and (real) 
utility, while $\ep$
combines the products to form the (generalized) expected utility,
in a way analogous to adding products of probabilities and utilities in
calculating standard expected utility.

We have
three domains because we do not want to require that
\dmakers\ be able to add or multiply plausibility values or utility values,
since these could be qualitative (\eg\ plausibility values could be
\commentout{
$\{$``unlikely'', ``likely'', 
``\mbox{very likely}'', 
$\ldots \}$ and utility values could be 
$\{$``bad'', ``good'', ``better'', $\ldots \}$)\@.
}%
\mbox{``unlikely''}, \mbox{``likely''}, \mbox{``very likely''}, etc.,
and utility values could be 
\mbox{``bad''}, \mbox{``good''}, \mbox{``better''}, etc.)\@.
In general, we do not assume that $\eleq$ is an order (or even a
\preorder), since we 
would like to be able to 
\represent\
as many 
\prefrels\ and 
\drules\ as possible.

Once we have \anexpstr, 
\dmakers\ can express their \tandb\ using 
components
of the \expstr.
More specifically, the \dmakers\
express their beliefs using a 
\emph{\plmsr}~\cite{FrH7}, whose range is the
\plstr\ of the \expstr\
(\plmsrs\ generalize probability 
measures and a host of other \representations\ of uncertainty, such as
sets of probability measures, Choquet capacities, possibility
measures, ranking functions, etc.)\  and they
express their tastes using a utility function whose range is 
the \utstr\ of the \expstr.
In \anexpstr, it is possible to define a generalization of expected
utility, which we call \emph{generalized \seu} (\gseu)\@. 
The \gseu\ of an act is basically the sum (in the sense of $\ep$) of products
(in the sense of $\et$) of plausibility values and utility values that
generalizes the standard definition of (probabilistic) expected utility over
the reals in the obvious way.

We start by proving 
an analogue
of Savage's 
result with respect to the \drule\ (Maximizing) \gseu\@.\footnote{Many
\drules\ involve optimizing 
(\ie\ maximizing or minimizing) 
some value function on the \acts.  
Sometimes it is explicitly mentioned
whether the function is to be maximized or minimized (\eg\ ``\reg'' says
explicitly to ``minimize the maximum regret'') while other times only the
function name is mentioned and it
is implicitly understood what is meant (\eg\ ``\seu'' means
``maximize \seu'')\@.  In this \paper\ 
we use ``\mgseu'' and ``\gseu'' interchangeably.}
\commentout{
We show that,
given 
any \prefrel\ 
on \acts\ (not necessarily an order), there is
some \expstr\ $\edom$, 
\plmsr\ $\plf$, and
\utfn\ $\utf$,
 such that \gseu\ 
with respect to
$\epu$
\represents\ the \prefrel;
}%
We show that every \prefrels\ on acts
has a \gseu\ \representation\ (even those that do not satisfy any of Savage's
postulates), where a \emph{\gseu\ \representation} 
of a \prefrel\ basically consists of an \expstr\ $\edom$, 
\plmsr\ $\plf$, and \utfn\ $\utf$, such that 
the way acts are related according to their \gseu\ agrees with the \prefrel\
\commentout{
that is, act $\act_1$ is preferred to act $\act_2$ 
according to the \prefrel\ iff the (generalized) expected
utility of $\act_1$ is higher than that of $\act_2$
}%
(\mthm{thm:anyorder}).
In other words,
no matter what the \dmaker's \prefrel\ on \acts\ is, 
she behaves
as if she has quantified her beliefs by a \plmsr\ 
and her tastes via a \utfn, 
and is 
\relating\ the
acts
according to their
(generalized) expected 
utility as defined by the $\ep$ and $\et$ of some expectation domain.
That is, 
we can model \emph{any} \dmaker\ using \gseu, whether
or not the \dmaker\ satisfies any 
\mbox{rationality}
assumptions.
\commentout{
An important difference between our result and that of
Savage is that we 
construct not only 
the plausibility measure and utility function, but also the expectation
domain.  
(For Savage, 
the \plstr\ is $[0,1]$ and both the \utstr\ and \valstr\ are $\realNum$,
the reals.)
Savage also shows that the probability measure is unique and the utility
function is unique up to affine transformations. We have no analogous
uniqueness results for our representation, although our choices are
canonical in a certain sense.
}%
An important difference between our result and that of Savage is that he was
constructing \emph{\seu} \representations, 
which consists of a \emph{real-valued} \utfn\ $\utf$
and a \emph{probability} measure $\Pr$ (and the \expstr\ is fixed, so $\ep$,
$\et$, and $\eleq$ are just
$+$, $\times$, and $\leq$, respectively).
\commentout{
\footnote{Also, as with most results,
Savage's 
result contains a uniqueness condition, which says, among other things, that 
there is a unique probability measure.
Thus his postulates are stronger than necessary for the \emph{existence} of
a \representation.  
Although we cannot show uniqueness, our choices are
canonical in a certain sense.
}
}%

Given that \gseu\ can \represent\ \emph{all} \prefrels, it might be
argued that \gseu\ 
is too general---it offers no 
guidelines 
as to how to make decisions.
We view this as a feature, not a bug, since 
our goal is to provide a general framework in which to express and study 
\drules, instead of proposing yet another \drule.
Thus the absence of ``guidelines'' is in fact an absence of 
\emph{limitations}\mbox{: we} do not want to exclude any possibilities at the
outset,  
even \prefrels\ that are not transitive or are \incomplete.
From the point of view of a behavioral scientist, this has the advantage
of allowing us to represent the preference relations 
that actually arise in real life, 
which  typically do not satisfy many of the standard
assumptions made by decision theorists, and doing so in a potentially
compact way (by specifying $\oplus$, $\otimes$, $\eleq$, a plausibility
measure $\plf$, and a utility function $\utf$).  Perhaps more
interesting is that, 
starting 
from a 
framework in which
we can \represent\ all \prefrels,
\commentout{
we can then 
impose constraints on \expstrs\ 
(as well as \utfns\ and \plmsrs)
so as to
restrict the \prefrels\ that can be \represented\  by \gseu\@.  
}%
we can then consider what \prefrels\ have ``special'' \representations, in the
sense that the \expstr, \plmsr, and \utfn\ in the \representation\ satisfy some
(joint) properties.
This allows us
to show how 
properties of \expstrs\ correspond to properties of \prefrels.
We can then ``customize'' \gseu\ by placing just the constraints we
want.  
We illustrate this by showing how each of Savage's postulates corresponds
in a precise sense to an axiom on \gseu\@.
This ability to customize \gseu\ may be of more interest to computer
scientists than to behavioral scientists.  If we try to design software
agents that make decisions on our behalf, it may not be appropriate to
assume that they will represent beliefs using probability measures
and tastes using real-valued utilities. For example, the information
that a system can obtain may be better modeled by a set of probability
measures than a single probability measure, and a user may represent his
or her tastes more qualitatively, using words like ``terrific'' and
``terrible'', rather than numerically.   Using \gseu, it should be
possible to design agents that make decisions based on more general
representations of beliefs and tastes, and customize them so that the
the decision-making process satisfies certain ``rationality'' postulates.

There is yet another advantage of this approach, which is the focus of
\cite{CH03b}.
Intuitively, a \drule\ 
maps 
tastes (and beliefs)
to 
\prefrels\ on acts.
Given two \drules\ $\dr_1$ and $\dr_2$, 
an \emph{$\dr_1$ \representation\ of $\dr_2$} is basically
a function $\tau$ that maps inputs of $\dr_2$ to inputs of $\dr_1$ that
\represent\ the same tastes and beliefs, with 
the property that $\dr_1(\tau(x)) = \dr_2(x)$.
Thus, $\tau$ models, in a precise sense, a user of $\dr_2$ as a user of
$\dr_1$, since $\tau$ preserves tastes (and beliefs).
In \cite{CH03b} 
\commentout{
we show that \gseu\ can represent, not
just all preference relations on acts, but also (almost) all decision
rules.  
}%
we show that many \drules\ have 
\gseu\ \representations.
Moreover, we show that (almost) every \drule\ has an 
\emph{ordinal} \gseu\ \representation,
where, rather than $x$ and $\tau(x)$ representing
exactly the same 
tastes (and beliefs), they preserve the 
relation between tastes (and beliefs),
without necessarily preserving the magnitude.  
For example,
if outcome $o_1$ is preferred to outcome $o_2$ in $x$, $o_1$ is also
preferred to $o_2$ in $\tau(x)$, although the magnitude of preference
may be different.

Our claim that only ``many'' \drules\ have a \gseu\
representation may seem inconsistent with our earlier claim that every
\prefrel\ has a \gseu\ representation.  Representing 
a \prefrel\ is not the same as representing a \drule.  If we again view a
\drule\ as a function from tastes (and possibly beliefs) to
\prefrel\ on alternatives, then \drule\ $\dr$ represents a
\prefrel\ $\preceq$
if there are some tastes and beliefs such that, with these as input,
$\dr$ returns $\preceq$.  On the other hand, $\dr_1$ represents
$\dr_2$ if, roughly speaking, for {\em all\/} possible inputs of
tastes (and beliefs), $\dr_1$ and $\dr_2$ return the same
\prefrel. That is, $\dr_1$ and $\dr_2$ act essentially the same
way as functions.
\commentout{
We refer the reader to
\cite{CH03b} 
for the formal definitions and details. 
(Formally, 
a \drule\ 
basically 
maps a \dprob---that is, 
a description of the DM's tastes and 
beliefs---to a 
\prefrel\ on acts.)

\commentout{
A \drule\ 
$\dr_1$ 
represents another \drule\ 
$\dr_2$ if there is a mapping $\tau$ on \dprobs\ 
that makes $\dr_1$ behave like $\dr_2$: \ie\ 
$\tau \circ \dr_1 = \dr_2$.  We refer the reader to
\cite{CH03b} 
for the formal definitions and details.  These results
provide further support for viewing \gseu\ as a universal \drule.
}%
}%

Although there has been a great deal of work on \drules,
there has been relatively little work on finding general frameworks for 
\representing\ \drules.  
In particular, there has been no attempt to find a \drule\ that
can represent all preference relations.
There has been work in the fuzzy logic
community on finding general notions of integration (which essentially
amounts to finding notions of expectation) using generalized notions of
$\ep$ and 
$\et$; 
see, for example,~\cite{BM00}.  However, 
the \expstr\ used in this work is (a subset of) the
reals; arbitrary \expstrs\ are not considered. 
Luce~\citeyear{Luce90,Luce00} 
also considers general 
addition-like operations applied to utilities,
but his goal is to model joint receipts.
Receipts are typically modeled in an arguably inefficient way, as
commodity bundles; it seems more natural to deal with them as Luce
does, in terms of an abstract binary operation, akin to the operations
we have here.

The rest of this paper is organized as follows.  We cover some basic
definitions in Section~\ref{sec:preliminary}\@: 
\plstrs, 
\utstrs, 
\expstrs,
\dprobs, and 
\gseu\@.
We show that 
every \prefrel\ on acts has a \gseu\ \representation\ 
in Section~\ref{sec:prefrels}.  In Section~\ref{sec:savage}, we show 
that each of Savage's postulates corresponds to an 
axiom on \gseu\@.  We conclude in Section~\ref{sec:discussion}.
\commentout{
Proofs of theorems stated are
available at http://www.cs.cornell.edu/home/halpern.

}%
Most proofs are deferred to the appendix.

\section{Preliminaries}\label{sec:preliminary}
\subsection{Plausibility, Utility, and Expectation Domains}

Since one of the goals of this \paper\ is to provide a general framework for
all of decision theory, we want to \represent\ the \tandb\ of the \dmakers\ in
as general a framework as possible.  
In particular, we do not want to force the \dmakers\ to 
linearly preorder all consequences and all events (\ie\ subsets of the set of states).
To this end, 
we use \plmsrs\ to
\represent\ the beliefs of the \dmakers\ and (generalized) \utfns\ to
\represent\ their tastes.

A \emph{\plstr} is a set $\pdom$, 
partially ordered by 
$\pleq_{\pdom}$  
(so $\pleq_{\pdom}$ is a reflexive, antisymmetric, and transitive
relation), 
with two 
special elements $\bot_{\pdom}$ and 
$\top_{\pdom}$, such that
(We often omit the subscript $\pdom$ in $\bot_\pdom$ and $\top_\pdom$
when it is clear from context.)
\commentout{
\mbox{for all $x \in \pdom$},
${\bottom_\pdom} \pleq_\pdom x \pleq_\pdom {\top_\pdom}$.
}%
${\bottom_\pdom} \pleq_\pdom x \pleq_\pdom {\top_\pdom}$ 
for all $x \in \pdom$.
Given a set $\Stt$, 
a function $\plf : 2^\Stt \rar \pldom$ is a \emph{\plmsr} iff
\ben
\renewcommand{\theenumi}{$\plf$\arabic{enumi}}
\settowidth{\itemindent}{\theenumi}
\bl
$\plf(\eset) = {\bottom}$, 
\bl
$\plf(\Stt) = {\top}$, and 
\bl
if $X \sbset Y$ then $\plf(X) \pleq \plf(Y)$.
\een

Clearly plausibility measures are 
generalizations
of probability
measures.  
As pointed out in~\cite{FrH7}, plausibility measures generalize a host
of other \representations\ of uncertainty as well.  Note that while the
probability of any two sets must be comparable (since $\realNum$ is
totally ordered), the plausibility of two sets may be incomparable.

We also want to \represent\ the tastes of \dmakers\ using something more
general than $\realNum$, so we allow the range of \utfns\ to be 
\utstrs, where a \emph{\utstr}
is a set $\udom$ endowed with a 
reflexive binary relation
$\eleq_{\udom}$.
Intuitively, elements of $\udom$ \represent\ the strength of likes
and dislikes of the \dmaker\ while elements of $\pdom$ \represent\ the
strength of her beliefs.
Note that we do not require the DM's preference to be transitive
(although we can certainly add this requirement).  Experimental evidence
shows that DM's preferences occasionally do seem to violate transitivity.

Once we have plausibility and utility, 
we want to combine them to form expected utility.  
To do this, we introduce \expstrs, which have utility domains,
 plausibility domains, and operators
$\ep$ 
(the analogue of $+$)
and $\et$ 
(the analogue of $\times$).\footnote{Sometimes we use $\times$ to denote
Cartesian product; the context will always make it clear whether this is the case.}
More formally, 
an \emph{\expstr} 
is a 
tuple 
$\edom = (\udom, \pldom, \eudom, \ep, \et)$, where 
$(\udom, \eleq_\udom)$ is a \utstr, $(\pldom,\pleq_\pldom)$ is a \plstr,
$(\eudom, \eleq_\eudom)$ is a \valstr\ (where $\eleq_\eudom$ is a
reflexive binary relation),  
$\et : \pldom \times \udom \rar \eudom$, and 
$\ep : \eudom \times \eudom \rar \eudom$.
There are four requirements on \expstrs:
\ben 
\renewcommand{\theenumi}{$\edom$\arabic{enumi}}
\settowidth{\itemindent}{\theenumi}
\setlength{\itemsep}{0in}
\bl 
\label{edom-eq-first}
\label{edom-eq1}
$(x \ep y) \ep z  =  x \ep (y \ep z)$;
\bl 
\label{edom-eq2}
$x \ep y  =  y \ep x$;
\bl
\label{edom-eq3}
${\top} \et x = x$;
\bl
\label{edom-eq-last}
\label{edom-eq4}
$(\udom, \eleq_{\udom})$ is a substructure of $(\eudom, \eleq_\eudom)$.
\een
\ref{edom-eq1} and \ref{edom-eq2} say that $\ep$ is associative and
commutative.  
\ref{edom-eq3} says that $\top$  is the left-identity of $\et$ and
\ref{edom-eq4} ensures that the \expstr\ respects the relation on
utility values.

Note that we do not require that $\ep$  be monotonic; that is, 
we do not require that
for all $x,y,z\in\eudom$, 
\beqn
\label{eqn:monotonic}
\mbox{if } x \eleq_\eudom y \mbox{ then } x \ep z \eleq_\eudom y \ep z.  
\eeqn
We say that $\edom$ is \emph{monotonic} iff 
\meqn{eqn:monotonic}
holds.
It turns out that
monotonicity does not really make a 
difference by itself;
\commentout{
see the comments after 
the proof of
\mthm{thm:anyorder}.
}%
see \mcor{cor:monotonic}.
\commentout{
Note that we also do not assume that $\bottom \et u$ is the identity for
$\ep$.  That is, we do not assume that
}%
Recall that in the standard case, $\bottom = 0$ and $0 \times x$ is the
identity for $+$.  In general, we do not assume that $\bottom \et u$ is the
identity $\ep$ (or that $\ep$ even has an identity).
We say that $\edom$ \emph{has a $\ep$ identity} iff
\beqn
\label{eqn:id}
(\bottom \et u) \ep x = x \mbox{ for all $u \in \udom$ and $x \in \eudom$.}
\eeqn
Because $\ep$ is commutative, there can clearly be
at most one identity for $\ep$, so if \meqn{eqn:id}
holds, then $\bottom \et u_1 = \bottom \et u_2$ for all  
$u_1,u_2 \in \udom$.
Requiring (\ref{eqn:id}) has very little effect on our results.  Some of
the proofs become easier, while others become somewhat more difficult,
but the theorems still hold.

Note that we also do not require $\et$ to distribute over $\ep$.  The
obvious way to state such a requirement is to require that $p \et (x \ep
y) = (p \et x) \ep (p \et y)$.  But this is not well-defined.  The
domain of $\et$ is $\pldom \times \udom$ and the domain of $\ep$ is 
$\eudom \times \eudom$.  If $x, y \in \udom$, then $x \ep y$, $p \et x$,
and $p \et y$ are all well defined, but $x \ep y$ may be an element of
$\eudom - \udom$, so $p \otimes (x \ep y)$ may not be well defined.
As we shall see, in cases where it the distributive property makes sense
(for example, if $\eudom = \udom$ or if $u_1 \ep u_2 \in \udom$ for
all $u_1, u_2 \in \udom$), then it actually does hold in many examples
of interest.

\xam\label{xam:expstr1}
The \emph{standard \expstr},
which we denote $\sedom$,
is 
$(\IR,[0,1],\IR,+,\times)$,
where the ordering on each domain is the standard order on the reals.
This, of course, is the \expstr\ which is used in defining
most \drules\ in the literature.  It is clearly monotonic and has a $+$
identity, namely 0.
\exam

\xam\label{xam:expstr2}
Consider the \expstr\ $\edom_2 = (\IR, [0,1] \times [0,1], \IR \times \IR,
\ep, \et)$, where 
\begin{itemize}
\item we use the standard order on the \utstr\ $\IR$;
\item the order $\pleq$ on the plausibility domain $[0,1] \times [0,1]$
is such that $(p_1,p_2) \pleq (q_1,q_2)$ iff $p_1 \le q_1$ and $p_2 \le
q_2$;
\item similarly, $(u_1,u_2) \eleq_{\eudom} (v_1,v_2)$ iff $u_1 \le
v_1$ and $u_2 \le v_2$;
\item $\ep$ is defined pointwise: $(u_1, u_2) \ep (v_1,v_2) = (u_1 + v_1,
u_2 + v_2)$;
\item $\et$ is pointwise multiplication: $(p_1,p_2) \et u = 
(p_1u, p_2 u)$.
\end{itemize}
We can view the \utstr\ $\IR$ as a substructure of the \valstr\ $\IR
\times \IR$ by identifying the element $u \in \IR$ with the pair
$(u,u)$.   Note that the ordering on the plausibility domain and the
ordering on the \utstr\ are both partial.
$\edom_2$ is also monotonic, and has $(0,0)$ as the
$\oplus$ identity.  
The distributive property (which makes sense here) is also easily seen
to hold: $(p_1,p_2) \et (u_1 \ep  u_2) = ((p_1,p_2) \et u_1) \ep
((p_1,p_2) \et u_2)$.

It turns out to also be of interest to consider the \expstr\ $\edom_2'$
which is defined just like $\edom_2$ except that the order
$\eleq_{\eudom}'$ on the \valstr\
is  defined by taking $(u_1,u_2) \eleq_{\eudom}' (v_1,v_2)$ iff
$\min(u_1,u_2) \le \min(v_1,v_2)$.  Note that this makes
$\eleq_{\eudom}'$ a total order.
\exam

\subsection{Decision Situations and Decision Problems}

A \emph{\dsitu} (under uncertainty) describes the objective part of the
circumstance 
that the 
\dmaker\ faces (\ie\ the part that is independent of the tastes and beliefs of
the \dmaker)\@.
We model a \dsitu\ in a standard way, as
a tuple $\ACT = (\Act, \Stt, \Csq)$, where 
\commentout{
\bit
\bl $\Stt$ is the set of \ssofn, and
\bl $\Csq$ is the set of \consequences.
\bl $\Act$ is a set of acts (\ie\ 
a set of 
functions from $\Stt$ to $\Csq$),
\eit
}%
\bit
\bl $\Stt$ is the set of \ssofn, 
\bl $\Csq$ is the set of \consequences, and
\bl $\Act$ is a set of acts (\ie\ 
a set of 
functions from $\Stt$ to $\Csq$).
\eit
An act $\act$ is \emph{simple} iff 
its range is finite.
That is, $\act$ is simple if it has only finitely many \consequences.
Many works in the literature focus on simple \acts\
(\eg\ \cite{Fishburn1987}).
We assume in this \paper\ that $\Act$ contains only simple acts; this means
that we can define (generalized) expectation using 
finite sums, so we do not have to introduce infinite series or integration for
arbitrary \expstrs.
Note that all  \acts\ are guaranteed to be simple if either $\Stt$ or
$\Csq$ is finite, although we do not assume that here.

A \dprob\ is essentially a \dsitu\ together with information about the \tandb\
of the \dmaker;  that is, a \dprob\ is a \dsitu\ together with the subjective
part of the circumstance that faces the \dmaker\@.
Formally,
a 
\emph{(\plstic) \dprob} is a tuple
$\dprb = (\ACT, \edom, \utf, \plf)$, where
\bit
\bl $\ACT = (\Act, \Stt, \Csq)$ is a decision situation,
\bl $\edom = (\udom,\pldom,\eudom,\ep,\et)$ is an \expstr, 
\bl $\utf : \Csq \rar \udom$ is a \utfn, and
\bl $\plf : 2^\Stt \rar \pdom$ is a \plmsr.
\eit
We say that $\dprb$ is \emph{monotonic} iff $\edom$ is monotonic.

\subsection{(Generalized) Expected Utility}
\label{sec:expectedutility:gseu}

Let $\dprb = ((\Act,\Stt,\Csq), \edom, \utf, \plf)$ be a \plstic\ \dprob.
Each $\act \in \Act$ induces a 
\emph{\urv}
$\utf_{\act} : \Stt \rar \udom$ 
as follows:
$\utf_{\act}(s) \bydef{=} \utf(\act(s))$.
In the standard setting (where utilities are real-valued and $\plf$
is a probability measure $\Pr$), we can identify the expected utility of
act $\act$ with the
expected value of $\utf_{\act}$
with respect to $\Pr$,
computed in the 
standard
way
(where we use $\ran(f)$ to denote the range of a function $f$):
\beqn
\label{eqn:std-meu}
\eprut{\act} \bydef{=} 
\sum_{x \in \ran(\utf_{\act})} \Pr(\utf_{\act}^{-1}(x)) 
\times x.%
\footnote{If the domain of $\Pr$ is some
nontrivial subalgebra of $2^\Stt$, then 
we must assume that $\utf_{\act}$ is a measurable function; that is,
$\utf_{\act}^{-1}(x)$ is a measurable set for all $x \in \ran(\utf_{\act})$.}
\eeqn
\commentout{
This expression for expected utility has an obvious generalization in an
arbitrary \expstr\ $\edom = (\udom, \pldom, \eudom, \et, \ep)$, 
}%
We can generalize \meqn{eqn:std-meu} to an arbitrary \expstr\ 
$\edom = (\udom, \pldom, \eudom, \ep, \et)$ by
\commentout{
where $+$ is 
replaced by $\ep$, 
$\times$ is replaced by $\et$, 
and $\Pr$ is
replaced by $\plf$.  
}%
replacing $+$, $\times$, and $\Pr$ by
$\ep$, $\et$, and $\plf$, respectively.
This gives us
\beqn
\label{eqn:def-E}
\eplut{\act} \bydef{=}
\EP_{x \in \ran(\utf_{\act})}
\plf(\utf_{\act}^{-1}(x)) \et x.
\eeqn
We call (\ref{eqn:def-E}) the
\emph{generalized \seu} (\gseu)
of act $\act$.
Clearly \meqn{eqn:std-meu} is a special case of \meqn{eqn:def-E}.

In the probabilistic case, if all singleton sets are measurable with
respect to $\Pr$ (i.e., in the domain of $\Pr$), then 
\begin{equation}
\label{eq0}
\eprut{\act} = \sum_{\stt \in \Stt} 
\Pr(\stt) \times \utf_{\act}(\stt).
\end{equation}
The plausibilistic analogue of (\ref{eq0}) is not necessarily equivalent to
(\ref{eqn:def-E}).  A \dprob\ $((\Act,\Stt,\Csq),\edom,\utf,\plf)$ is
\emph{additive} iff, 
for all $c \in \Csq$ and nonempty $X,Y \sbset \Stt$
such that $X \cap Y = \eset$,  
\[
\plf(X \cup Y) \et \utf(\csq) = 
(\plf(X) \et \utf(\csq)) \ep (\plf(Y) \et \utf(\csq)).
\]
Note that the
notion of additivity we defined is a joint property of several components of a
\dprob\ (\ie\ $\ep$, $\et$, $\utf$, and $\plf$) instead of being a property of
$\plf$ alone.  
Additivity is exactly the requirement needed to make the analogue of
(\ref{eq0}) equivalent to (\ref{eqn:def-E}).  While 
\dprobs\ involving probability are additive, those involving
representations of uncertainty such as Dempster-Shafer belief functions
or,  more generally, Choquet capacities,
are not, in general.  

\xam\label{xam:eu1}
For a \dprob\ $(\ACT, \sedom, \utf, \Pr)$, where $\sedom$ is the
standard \expstr\ and $\utf$ is a real-valued \utfn, \gseu\ agrees with
\seu.
\exam

\xam\label{xam:eu2} Consider the \dprob $(\ACT,\edom_2,\utf, (\Pr_1,\Pr_2))$,
where $\edom_2$ is the \expstr\ described in  
Example~\ref{xam:expstr2} and $\utf$ is a real-valued \utfn.  The pair
$(\Pr_1,\Pr_2)$ of probability measures can be viewed as  a single
plausibility measure.  If $\ACT = (\Act, \Stt,\Csq)$, then the
plausibility of $X \subseteq \Stt$ is a pair $(\Pr_1(X), \Pr_2(X))$.  It
is easy to check that 
$$\E_{(\Pr_1,\Pr_2),\edom_2}(\utf_{\act}) = (\E_{\Pr_1}(\utf_{\act}),
\E_{\Pr_2}(\utf_{\act}).$$  
Moreover, 
$\E_{(\Pr_1,\Pr_2),\edom_2}(\utf_{\act}) \eleq_{\eudom}
\E_{(\Pr_1,\Pr_2),\edom_2}(\utf_{\act'})$ iff
$\E_{\Pr_i}(\utf_{\act}) \le
\E_{\Pr_i}(\utf_{\act'})$ for $i = 1,2$.

On the other hand, if we consider $\edom_2'$, we still have 
$\E_{(\Pr_1,\Pr_2),\edom_2'}(\utf_{\act}) = (\E_{\Pr_1}(\utf_{\act}),
\E_{\Pr_2}(\utf_{\act})$, but now 
$\E_{(\Pr_1,\Pr_2),\edom_2'}(\utf_{\act})
\E_{(\Pr_1,\Pr_2),\edom_2'}(\utf_{\act'})$ iff
$\min(\E_{\Pr_1}(\utf_{\act}, \E_{\Pr_2}(\utf_{\act}))
 \le \min(\E_{\Pr_1}(\utf_{\act'}, \E_{\Pr_2}(\utf_{\act'}))$.

We can think of the plausibility measure $(\Pr_1,\Pr_2)$ as describing a
situation where the DM is unsure which of $\Pr_1$ and $\Pr_2$ is the 
``right'' probability measure.  In $\edom_2$, act $\act$ is considered
at least as good as $\act'$ if it is at least as good no matter which of
$\Pr_1$ and $\Pr_2$ describes the actual situation.  In $\edom_2'$,
$\act$ is at least as good as $\act'$ if the worst expected outcome of
$\act$ (with respect to each of $\Pr_1$ and $\Pr_2$) is at least as good
as the worst expected outcome of $\act'$.
Note how relatively different orderings of \valstr\ can
produce quite different ordering on acts, using \gseu.
\exam

Another example of the use of \gseu\ can be found in the proof of
Theorem~\ref{thm:anyorder}.  Although the construction does not
correspond to any standard \dprob\ in the literature, it does show
how
flexible the approach is.

\section{\Representing\ Arbitrary Preference Relations}
\label{sec:prefrels}

In this section, we show that 
every \prefrel\ on acts has a \gseu\ \representation.
\commentout{
To make the notion of representation precise, note that 
\gseu\ can be viewed as a function from decision problems to 
relations on acts.  
}%
\gseu, like all \drules, is formally a function from \dprobs\ to \prefrels\ on
acts. 
Thus a \gseu\ \representation\ of a \prefrel\ $\leqact$ on the acts in 
$\ACT = (\Act,\ldots)$ is a \dprob\ $\dprb = (\ACT,\edom,\utf,\plf)$,
where $\edom = (\udom,\pdom,\eudom,\ep,\et)$, such that
\commentout{
$\mr{\gseu}(\dprb)$
is the preference relation
$\precsim$ such that 
$\act_1 \eleq \act_2$ 
}%
$\act_1 \leqact \act_2$
iff $\eplut{\act_1} \eleq_\eudom \eplut{\act_2}$.
\commentout{
\gseu\ \represents\ a \prefrel\ $\leqact$ on the acts in 
$\ACT = (\Act,\ldots)$ 
if there exists an \expstr\ $\edom$, a
\plmsr\ $\plf$, and a \utfn\ $\utf$, such that $\act_1 \leqact \act_2$ iff
$\eplut{\act_1} \eleq_\edom \eplut{\act_2}$.  In other words, 
\gseu\ \emph{\represents} $\leqact$ on the acts in $\ACT = (\Act,\ldots)$ iff
there exists some 
decision problem
$\dprb = (\ACT,\ldots)$ such that 
$\mr{\gseu}(\dprb) = {\leqact}$.
}%

\thm\label{thm:anyorder} 
\commentout{
Given a preference relation 
$\leqact$ on the
acts in 
a decision situation 
$\ACT = (\Act, \ldots)$, 
there is an additive decision problem
$\dprb = (\ACT, \ldots)$ such that 
$\mr{\gseu}(\dprb) = {\leqact}$. 
}%
Every 
preference relation
$\leqact$ has a \gseu\ \representation.
\ethm

\prf
Fix some $\ACT = (\Act, \Stt, \Csq)$ and $\leqact$.
We want to construct a \dprob\ $\dprb = (\ACT, \edom, \utf, \plf)$ 
such that $\mr{\gseu}(\dprb) = {\leqact}$.  

The idea is to let each consequence be its own utility and each set be its own
plausibility, and define $\et$ and $\ep$ such that each act is its own expected
utility. 
For each $\csq \in \Csq$, let $\act_\csq$ denote the constant act
with the property that $\act_\csq(\stt) = \csq$ for all $\stt \in \Stt$.
Let $\edom = (\udom, \pdom, \eudom, \ep, \et)$ be defined as 
follows:
\begin{enumerate}
\item 
$\udom = (\Csq, \eleq_\Csq)$, where $\csq \eleq_\Csq \csqd$ iff
$\csq = \csqd$ or $\act_\csq, \act_\csqd \in \Act$ and 
$\act_\csq \leqact \act_\csqd$. 
(Note that Savage assumes that $\Act$ contains all simple acts; in particular,
$\Act$ contains all constant acts. We do not assume that here.)
\item $\pdom = (2^\Stt, \sbset)$.
\item 
$\eudom = (2^{\Stt \times \Csq}, \eleq_\eudom)$, where
$x \eleq_\eudom y$ iff $x = y$ or $x, y \in \Act$ and $x \leqact y$.  (Note
that set-theoretically a function is a set of 
ordered pairs, so $\Act \sbset 2^{\Stt \times \Csq}$.)
\commentout{
Note that since a function can be 
identified with a set of ordered pairs, acts in $\Act$ can be viewed as
elements of $\eudom$.  Take $x \eleq_\eudom y$ iff $x = y$ or $x = \act$  and
$y = \actb$ for $\act, \actb \in \Act$ such that $\act \leqact \actb$.
}%
\item $x \ep y = x \cup y$ for $x, y \in \eudom$.
\item $X \et \csq = X \times \{\csq\}$ for $X \in 2^\Stt$  $(= \pdom)$ and 
$\csq \in \Csq$ $(= \udom)$.
\end{enumerate}

We can identify 
$\csq \in \Csq$ with $\Stt \times \{\csq\}$ in $\eudom$; 
with this identification, $(\udom,\eleq_\udom)$ is a substructure of
$(\eudom, \eleq_\eudom)$ and $\top \et c = c$ for all $c \in \udom$ $(= \Csq)$,
as required. 
Furthermore, $\ep$ is clearly associative and commutative, so $\edom$ is indeed
an \expstr. 
Let $\dprb = (\ACT, \edom, \utf, \plf)$, where $\utf(\csq) = \csq$ and
$\plf(X) = X$. 
\commentout{
It is easy to check that $\dprb$ is additive, since if $X \cap Y = \eset$, then
\[
(X \cup Y) \times \{\csq\} = (X \times \{\csq\}) \cup (Y \times \{\csq\}).
\]
}%
Note that 
\[
\begin{array}{lcl}
\eplut{\act} &
= & 
\ds 
{\EP}_{x \in \ran(\utf_{\act})}
\plf(\utf_{\act}^{-1}(x)) \et x \\
& = & 
\ds 
{\EP}_{\csq \in \ran(\act)} \plf(\act^{-1}(\csq)) \et \csq \\
& = & 
\ds 
\{ (\stt, \csq) \st \act(\stt) = \csq \} 
\\
& = & 
\ds 
\act.
\end{array}
\]
That is, each act is its own expected utility;
by the definition of $\eleq_\eudom$, 
it is clear that 
$\act \leqact \actb$ iff $\eplut{\act} \eleq_\eudom \eplut{\actb}$. 
Thus $\mr{\gseu}(\dprb) = {\leqact}$, as desired.
\eprf
\commentout{
Note that, unlike most
\representation\ theorems, there is no uniqueness condition in
Theorem~\ref{thm:anyorder}.  
However, the \representation\ constructed in \mthm{thm:anyorder} is 
canonical in the following sense.  Suppose that $\dprb' = (\ACT, \edom',
\utf', \plf')$ is an arbitrary  \gseu\ \representation of $\leqact$, where 
$\edom' = (\udom',\pdom',\eudom',\ep',\et')$ and $\ACT = (\Act,
\Stt,\Csq)$.  Let $\udom'' = \{\utf'(c): c \in \Csq\}$, let
$\pdom'' = \{\plf'(X): X \subseteq \Stt\}$, and let $\eudom''$ be the
subset of $\eudom'$ consisting of all elements of the form 
$\plf'(X_1) \et' \utf'(c_1)) \ep' \cdots \ep' (\plf'(X_k) \et'
\utf'(c_k)$, where $X_1, \ldots, X_k$ are pairwise disjoint subsets of
$\Stt$.  Finally, let $\dprb = (\ACT, \edom,
\utf, \plf)$ be the \gseu\ \representation\ of $\leqact$ constructed in 
\mthm{thm:anyorder}.  Then there is an embedding of $(\udom'',\pdom'',
\eudom'', \ep', et')$ into $(\udom,\pdom,\eudom, \ep, \et)$, that is, an
injection $f$ mapping 
$\udom''$ to $\udom$, $\pdom''$ to $\pdom$, and $\eudom''$ to 
$\eudom$ such that 
\begin{itemize}
\item $f(x \et'  y) = f(x) \et f(y)$ for all $x \in
\pdom''$, $y \in \udom''$, and
\item $f(x \ep' y) = f(x)  \ep f(y)$ for all $x, y \in \eudom''$ such
that $x \ep' y \in \eudom''$.
\end{itemize}
}%
Note that, unlike most
\representation\ theorems, there is no uniqueness condition in
Theorem~\ref{thm:anyorder}.  
This is because, unlike most \representation\ theorems,
we do
not assume that the \expstr\ is $\sedom$, the standard \expstr, and we
do not 
assume that $\leqact$ \satisfies\ any assumptions.  
So one reason for 
the lack of uniqueness in \mthm{thm:anyorder} is because we place no
restriction whatsoever on $\leqact$. 
\commentout{
For example,
suppose that $\Act = \{\act_1,\act_2\}$, $\Stt = \{\stt_1, \stt_2\}$, 
$\Csq = \{\csq_1, \csq_2\}$, and $\act_1 \lsact \act_2$, then it is not
hard to show that there are infinitely many probability measures and
utility functions that will \represent\ $\leqact$.
}%
The other reason for the lack of uniqueness is that we consider arbitrary
\expstrs\ instead of restricting ourselves to $\sedom$.
Note that, 
even if $\leqact$ satisfies all of Savage's postulates,
although there is 
a unique \gseu\ \representation\ of $\leqact$ using
the standard \expstr\ $\sedom$ 
and probability measures (this is essentially Savage's result), 
there is no 
unique 
\gseu\ \representation\ 
if we allow arbitrary \expstrs.
\commentout{
Moreover, if we drop some of Savage's postulates, then there is no
longer a unique representation, even if we restrict to the standard
expectation domain.
}%
In particular,
the \representation\ constructed in the
proof of \mthm{thm:anyorder} is certainly distinct from the one
Savage~\citeyear{Savage} constructs. 
\commentout{
Given that we do not make
these standard assumptions, two questions suggest themselves:  Fix a \drule\
$\dr$ (say \eu\ or \gseu), 
\ben
\bl 
Given a set of joint properties of \expstrs, \utfn, and \plmsrs, is there a set of
assumptions
on \dsitus\ and \prefrels\ such that, $\ACT = (\Act,\Stt,\Csq)$
and $\leqact$ satisfy the assumptions iff there exists a unique \plstic\
\dprob\ $\dprb = (\ACT,\edom,\utf,\plf)$ such that 
$\dr(\dprb) = {\leqact}$ and $\edom$, $\utf$, and $\plf$ satisfy the
joint properties?  

\bl
Given a set of assumptions on \dsitu\ and \prefrels, is there a set of
joint properties 
of \expstrs, \utfn, and \plmsrs\ such that,  $\ACT = (\Act,\Stt,\Csq)$
and $\leqact$ satisfy the assumptions iff there exists a unique \plstic\
\dprob\ $\dprb = (\ACT,\edom,\utf,\plf)$ such that 
$\dr(\dprb) = {\leqact}$ and $\edom$, $\utf$, and $\plf$ satisfy the
joint properties?  

\een

Most results in the decision-theory literature focus on the first question, with the
joint properties being that the \expstr\ is $\sedom$, the \utfn\ is
real-valued, and the \plmsr\ satisfies some axioms (say those of probability
measures or those of belief functions, etc.)\@.  For example, Savage
characterizes when it is possible to \represent\ a \prefrel\ uniquely using
\seu~\cite{Savage}.
The second question does not typically arise in the
decision-theory literature, since the \expstr\ considered is usually the
standard one.  
Recall that whether the equation $x^3 = 1$ has a unique solution
depends on whether we are considering the real numbers or the complex numbers (or some
other ring or field).  In a sense, a \representation\ of a \prefrel\ can be
viewed as a ``solution'' to a system (that depends on the \drule, the \expstr,
this is especially true in the standard 
case), and unique \representability\ essentially means unique solvability.
Thus whether a \prefrel\ (that may or may not satisfy some
assumptions) has a unique  \representation\ depends on what \expstrs, \utfns,
and \plmsrs\ we consider.  
}%
While  there is no unique \gseu\ \representation, the \gseu\
\representation\ we 
construct in the proof of \mthm{thm:anyorder} is 
canonical in the following sense. 
Fix a \dsitu\ $\ACT = (\Act,\Stt,\Csq)$ and a
\prefrel\ $\leqact$.  
Suppose $\dprb = (\ACT,\edom,\utf,\plf)$ is the \dprob\
constructed in the proof of \mthm{thm:anyorder} and let 
$\dprb_0 = (\ACT,\edom_0,\utf_0,\plf_0)$ be an arbitrary \gseu\
\representation\ 
of $\leqact$, where 
$\edom_0 = (\udom_0,\pdom_0,\eudom_0,\widehat{\ep},\widehat{\et})$.
It is easy to check that 
\bit
\bl
for all $X, Y \sbset \Stt$, 
$\plf(X) \pleq_\pdom \plf(Y)$ implies $\plf_0(X) \pleq_{\pdom_0} \plf_0(Y)$, 
\bl
for all $\csq, \csqd \in \Csq$, 
$\utf(\csq) \eleq_{\udom} \utf(\csqd)$ implies
$\utf_0(\csq) \eleq_{\udom_0} \utf_0(\csqd)$, and
\bl
for all $X_1, \ldots, X_n, Y_1, \ldots, Y_m \sbset \Stt$,
for all $\csq_1, \ldots, \csq_n, \csqd_1, \ldots, \csqd_m \in \Csq$, 
\[
\plf(X_1) \et \utf(\csq_1) \ep \cdots \ep \plf(X_n) \et \utf(\csq_n) \eleq_{\eudom}
\plf(Y_1) \et \utf(\csqd_1) \ep \cdots \ep \plf(Y_m) \et \utf(\csqd_m) 
\]
implies
\[
\plf_0(X_1) \widehat{\et} \utf_0(\csq_1) \widehat{\ep} \cdots \widehat{\ep} \plf_0(X_n) \et \utf_0(\csq_n) \eleq_{\eudom_0}
\plf_0(Y_1) \widehat{\et} \utf_0(\csqd_1) \widehat{\ep} \cdots \widehat{\ep} \plf_0(Y_m) \et \utf_0(\csqd_m).
\]
\eit
Thus, the \representation\  we construct is minimal, in the sense that
we relate only what has to be related to satisfy the definition of \representation.
\commentout{
More precisely, given a \plstic\ \dprob\ 
$\dprb = ((\Act,\Stt,\Csq),\edom_1,\utf_1,\plf_1)$, w
\bit
\bl
${\eleq_\Csq^{\dprb}} = \{ (\csq, \csqd) \st \utf(\csq) \eleq_{\udom} \utf(\csqd) \}$,
\bl
${\eleq_\Stt^{\dprb}} = \{ (X, Y) \st \plf(X) \pleq_{\pdom} \plf(Y) \}$, and
\bl
${\eleq_{\eudom}^{\dprb}} = \{ (\plf(X_1) \et \utf(\csq_1) \ep \cdots \ep
\plf(X_n) \et \utf(\csq_n), \plf(Y_1) \et \utf(\csqd_1) \ep \cdots \ep
\plf(Y_m) \et \utf(\csqd_m)) \st 
\plf(X_1) \et \utf(\csq_1) \ep \cdots \ep \plf(X_n) \et \utf(\csq_n) \eleq_{\eudom}
\plf(Y_1) \et \utf(\csqd_1) \ep \cdots \ep \plf(Y_m) \et \utf(\csqd_m)\}$.
\eit 
}%
Moreover,
the \representation\ constructed in \mthm{thm:anyorder} is in
fact additive, since if $X \cap Y = \eset$, then 
$(X \cup Y) \times \{\csq\} = (X \times \{\csq\}) \cup (Y \times
\{\csq\})$,
and has a $\ep$ identity, namely $\emptyset$.
\commentout{
The \representation\ is not necessarily monotonic.  
However, as we show in the
full paper, a simple modification to the definition of $\eleq_\eudom$ makes the
\representation\ monotonic, so monotonicity by itself does not restrict the
kind of \prefrels\ that can be \represented.
}%

The \expstr\ $\edom$ constructed in the proof of \mthm{thm:anyorder} is not
(necessarily) \epmono, since we certainly could have two acts $\act$ and
$\actb$ such that $\act \leqact \actb$, 
so $\eplut{\act} \eleq_\eudom \eplut{\actb}$, but there is some $x \in \eudom$ such
that $\eplut{\act} \ep x \not\eleq_\eudom \eplut{\actb} \ep x$.  
In fact, our construction has the property that
two distinct expressions are unrelated unless
they are both expected utility values.
As the following corollary shows,
it is not  hard to 
modify the proof by extending 
$\eleq_\eudom$
so as to 
make $\edom$ \epmono.
\newcounter{monoSecNum}
\newcounter{monoThmNum}
\setcounter{monoSecNum}{\value{section}}
\setcounter{monoThmNum}{\value{THEOREM}}
\cor
\label{cor:monotonic}
Every \prefrel\ has a \epmono\ \dpadd\ \gseu\ \representation\
with a $\ep$ identity.
\ecor
\prf
See the appendix. \eprf

\mcor{cor:monotonic} shows 
that requirements like monotonicity,
additivity, and having 
a $\ep$ identity do not
restrict the kind of 
\prefrel\ that \gseu\ can \represent.  
But this means that these requirements do not by themselves
prevent \gseu\ from producing ``strange'' \prefrels\ when
it is applied as a \drule.  
In the next section, we consider constraints on expectations domains
that do force the \prefrel\ produced by \gseu\ to be arguably more reasonable.

\commentout{
Theorem~\ref{thm:anyorder} can be viewed as an analogue  of Savage's
result.  Savage 
showed that any preference relation
on acts
satisfying certain postulates can be
represented using standard expected utility; Theorem~\ref{thm:anyorder}
shows that any preference relation on acts can be represented using the greater
flexibility allowed by applying expectation to an arbitrary additive decision
problem.  (In fact, we can prove Theorem~\ref{thm:anyorder} even if we
put further restrictions on the decision problem, such as requiring
$\ep$ to be monotonic.)
}%

Theorem~\ref{thm:anyorder} holds in large part because of the flexibility
we have.  Given a \dsitu\ $(\Act,\Stt,\Csq)$
and a \prefrel\ $\leqact$ on $\Act$,
we are able to construct an \expstr\ and a relation $\eleq_\eudom$ that is
customized
to capture the relation $\leqact$ on $\Act$.
\commentout{
I noted this before, around 10/2001:

----
In section 3, however, we were constructing an expectation structure for
each preference order.  (So we are definitely not using a single domain: 
it seems like a cheat.) There is more than one way to fix this in the new
framework.  We could construct a single expectation structure for each
decision situation (see below and the writeup) or a single expectation
====

I also brought this up in some meeting(s), but I think the comment was that
this has nothing to do with decision theory: it's just some encoding trick.

FWIW, if we are going to include this, I'd like to point out that we need much
less freedom than even what you wrote.  I claim that we can construct
expectation domains without any reference to any specific decision situation.  
Perhaps the simplest way to illustrate this is to consider an ``abstract'' decision
situation in which the set of consequences is some cardinal and the set of
states is some cardinal.  Then do the construction you did (which is basically
what I had in mind when I brought up this issue).  For each pair of cardinals
(K,L), we have an expectation structure E_{K,L}.  We could even let E_{K,L}
also contain all E_{K',L'} for K' \leq K and L' \leq L as ``subdomains'', so 
E_{K,L} can handle any decision situation whose |S| and |C| are bounded above
by K and L, respectively.  Perhaps we can simplify things further by indexing 
the expectation structure using a single cardinal E_{K} = E_{K,K} (so we bound
the size both sets using the same cardinal).  Then what we have is a sequence
of expectation domains E_{K} for all cardinals K > 0 such that for any decision
situation and prefrel \le, we can find some E_K (basically K = max(|S|,|C|)) 
such that we can represent \le using some utfn and plmsr.  This is more
analogous to Savage's result, since we no longer construct an expstr that is
``tailor-made'' for the prefrel:  we have, once and for all, a sequence of
expectation domains that works (basically, these contain all possible reflexive
binary relations of interest and we just have to pick them out using the
utility function).

And, of course, the only reason we don't have
a *single* expectation domain is because of cardinality considerations; E_K
will, however, work for all decision situations such that |S|,|C| \le K -- so
there is a single domain once the size of the sets are bounded.  Furthermore,
in what I have in mind, E_K is a substructure of E_K' for K \leq K', so the
``limit'' of the sequence is an ``expectation domain'' (which would be a proper
class, like the collection of all ordinals), that would work for all set-valued
decision situations.  

I'll let you decide what you want to do with this.  A footnote perhaps?
}%
We do not need quite this much flexibility.  We can strengthen 
Theorem~\ref{thm:anyorder} to show that 
for every decision situation $\ACT = (\Act, \Stt, \Csq)$,
there exists an \expstr\  $\edom_{\ACT}$ such that for all preference
relations $\leqact$ on $\Act$, there exists a utility function $\utf$
and plausibility measure $\plf$ such that
$\mr{\gseu}((\ACT,\edom,\utf,\plf) = {\leqact}$.  
That is, given $\ACT$,
we can fix the 
\expstr\ 
once and for all, rather than
taking a different expectation structure (more precisely, a different
order $\eleq_\eudom$ on the \valstr) for each preference relation
$\leqact$.  Indeed, we can even fix the plausibility measure once and
for all as well.

\newcounter{aoSecNum}
\newcounter{aoThmNum}
\setcounter{aoSecNum}{\value{section}}
\setcounter{aoThmNum}{\value{THEOREM}}
\thm\label{thm:anyorder'} Given a decision situation 
$\ACT = (\Act,\Stt,\Csq)$, there 
exists a monotonic, additive \expstr\ $\edom$ and a
plausibility measure $\plf$ on $\Stt$ such that,
for every preference
relation $\leqact$ on $\Act$, there exists a utility function
$\utf_{\leqact}$ on 
$\Csq$ and that 
${\leqact} = \mr{\gseu}(\dprb)$, 
where $\dprb = (\ACT,\edom,\utf_{\leqact},\plf)$.
\ethm

\prf 
Again, the argument proceeds by modifying the construction in
Theorem~\ref{thm:anyorder}.  We leave details to the appendix.
\eprf

Theorems~\ref{thm:anyorder} 
and~\ref{thm:anyorder'} 
depend 
(in part) on two features of our setup.  
The first
is that, following Savage
\citeyear{Savage}, 
we took \acts\ to be functions from \states\ to 
\consequences.
This is not an entirely trivial assumption. In
practice, 
different acts might produce the same consequences,
depending on how the consequences are modeled.
For 
example, suppose that Alice has a red umbrella and a blue umbrella (both in
good condition).  If the set of consequences is $\{\mbox{``getting wet''}, \mbox{``staying dry''}\}$,
then carrying the red umbrella will produce the same \consequences\ as carrying
the blue umbrella.
Suppose instead
that we have a \emph{\consequence\ function}
$\csqf : \Act \times \Stt \rar \Csq$
that takes an act $\act$ and a state $\stt$ and gives the \consequence\ of
$\act$ in $\stt$.  
Of course, in this setting, two 
distinct
\acts\ $\act_1$ and $\act_2$ could induce the same function from 
\states\ to \consequences; that is, 
we might have $\csqf(\act_1, \stt) = \csqf(\act_2, \stt)$ for all $\stt
\in \Stt$. 
It is easy to 
see
that if $\act_1$ and $\act_2$ induce the same function
from \states\ to \consequences, then 
no matter what \expstr, \utfn, and \plmsr\ we use, $\act_1$ and $\act_2$
will have the 
same expected utility.
Thus,  
if $\leqact$ does not treat $\act_1$ and $\act_2$ the same way, 
then $\leqact$ 
\commentout{
cannot be represented by
\gseu\@.
}%
has no \gseu\ \representation.  
(An analogue of Theorem~\ref{thm:anyorder} holds in this 
\mbox{case: as}
long
as two acts that induce the same function are treated the same way by
$\leqact$, then $\leqact$ has a \gseu\ \representation.)

A second reason that we do not need consistency constraints 
on $\leqact$
is that 
we have placed no constraints on $\eleq_\eudom$, and relatively few
constraints on $\ep$, $\et$, $\utf$, and $\plf$.  
If,
for example, we 
required $\eleq_\eudom$ to be transitive, 
then we would also have to require that $\leqact$ be transitive.
The lack of constraints on $\ep$, $\et$, $\utf$, and $\plf$ is important
because it gives us enough freedom to ensure that 
distinct acts have different expected utility.
In 
the next section, 
we investigate what happens when we add more constraints.

\section{Representing Savage's Postulates}
\label{sec:savage}

\mthm{thm:anyorder} shows that \gseu\ can \represent\ any \prefrel.
We are typically interested in 
\representing\ 
\prefrels\ that satisfy
certain constraints, or postulates.  
The goal of this section is to examine the effect of such constraints on
the components that make up \gseu\@.
For definiteness, we focus on Savage's postulates. 
\commentout{
For ease of exposition, we restrict to additive decision problems in
this section; this assumption simplifies some of the more technical
arguments.  Recall that this restriction does not affect
Theorems~\ref{thm:anyorder} and \ref{thm:anyorder'}.
}%

A set $\dm_e$ of axioms about 
(\ie\ constraints on) 
\plstic\ \dprobs\
\emph{\represents} a set of
postulates 
$\dm_r$
about
\dsitu\ and \prefrel\ pairs \emph{with respect to a collection of \dprobs\ $\dprbs$}
iff
for all $\dprb \in \dprbs$, 
\commentout{
\bit
\bl
$\BB{\dprb} = (\ACT, \edom, \utf, \plf)$ \satisfies\ $\dm_e$ iff
$(\ACT,\mr{\gseu}(\dprb))$ \satisfies\ $\dm_r$.
\eit
}%
\[
\BB{\dprb} = (\ACT, \edom, \utf, \plf) \mbox{ \satisfies\ } \dm_e \mbox{ iff }
(\ACT,\mr{\gseu}(\dprb)) \mbox{ \satisfies\ } \dm_r.
\]
\mthm{thm:anyorder} can be viewed as saying that the empty set of axioms
\represents\ the empty set of postulates
with respect to the collection of all \plstic\ \dprobs.
\commentout{
Note that if $\dm_e$ \represents\
then $\dm_e$ \represents\ $\dm_r$ with respect to $\dprb_1$.
}%
Note that if $\dm_e$ \represents\
$\dm_r$ with respect to $\dprbs_0$  and $\dprbs_1 \sbset \dprbs_0$,
then $\dm_e$ \represents\ $\dm_r$ with respect to 
$\dprb_1$ as well.

Before we present Savage's postulates, we first introduce some notation that
will make the exposition more succinct.
Suppose that $f : X \rar Y$, $g : X \rar Y$, and $Z \sbset X$.  Let
$\lotto{f}{Z}{g}$ denote the function $h$ such that $h(x) = f(x)$ for
all $x \in Z$ and $h(x) = g(x)$ for all $x \in \comp{Z}$.  
For example, if $X = Y = \realNum$ and 
$Z = \{x \st x < 0\}$, then 
$\lotto{-x}{Z}{x}$ 
is the absolute value function.
In the intended application, the functions in question will be acts (\ie\
functions from the set of states $\Stt$ to \consequences\ $\Csq$). So
$\act = \lotto{\act_1}{X}{\act_2}$ is the act such that 
$\act(\stt) = \act_1(\stt)$ for all $\stt \in X$ and 
$\act(\stt) = \act_2(\stt)$ for all $\stt \in \comp{X}$.  For brevity, 
we identify the consequence $\csq \in \Csq$ with the constant act
$\act_{\csq}$ such that $\act_{\csq}(\stt) = \csq$ for
all $\stt \in \Stt$.  So for $\csq_1, \csq_2 \in \Csq$,
$\lotto{\csq_1}{X}{\csq_2}$ is the act with the property that
$\act(\stt) = \csq_1$ for all $\stt \in X$ and 
$\act(\stt) = \csq_2$ for all $\stt \in \comp{X}$.
Recall that 
$X_1, \ldots, X_n$ 
is a \emph{partition} of $Y$ iff
the $X_i$'s are nonempty and pairwise disjoint, and $\bigcup_i X_i = Y$.

Fix some \dsitu\ $(\Act,\Stt,\Csq)$.
Readers familiar with~\cite{Savage} will recall that Savage 
assumes
that $\Act$ consists of all possible functions from $\Stt$ to $\Csq$, since the
\dmaker\ can be questioned about any pair of functions. 
(Though when Savage
proves the main theorem in Chapter~5~of~\cite{Savage}, he restricts attention
to acts that induce simple lotteries, since he essentially reduces his problem
to the one already solved by \vonNM~\citeyear{vNM1947}, and \vonNM\
focused on 
simple lotteries.)  
This is a rather strong assumption.  It means that the DM is required to have
preferences on a rather large set of acts, many of which are not in his
power to perform (and, indeed, many of which might be impossible to
realize).  
Savage needs this assumption for his theorem.
We do not need it for our results, although making this assumption
simplifies the statement of the relevant axioms.
As we have throughout this paper, in this section, we continue to allow
$\Act$ to
be any nonempty subset of the set of all simple acts.
\commentout{
It is possible to maintain that assumption here, though some of the postulates
would fail, not because $\leqact$ does not relate certain members of
$\Act$,  but
because $\Act$ does not contain certain pairs of acts.
We could adapt the postulates by ``relativizing'' them (so
acts not in $\Act$ are not required to be related) as we do in the full paper;
however, that involves changing the
statements of Savage's postulates somewhat, and the presentation becomes more 
complicated.
To simplify the exposition here,
we assume in this section only that $\Act$ consists of \emph{all} simple
acts.  
Here are Savage's first six postulates 
(stated under the assumption that $\Act$ consists of all simple acts
from $\Stt$ to $\Csq$,
so that, for example, if $a$ and $b$ are in $\Act$, and $X \subseteq S$,
then $(a,X,b) \in \Act$):

\ben
\renewcommand{\theenumi}{P1}
\bl
\label{sq:p1}
For all $\act_1, \act_2, \act_3 \in \Act$, 
\bit
\item[(a)] $\act_1 \leqact \act_2$ or 
$\act_2 \leqact \act_1$, and
\item[(b)]
if $\act_1 \leqact \act_2$ and $\act_2 \leqact \act_3$, then 
$\act_1 \leqact \act_3$.
\eit
\renewcommand{\theenumi}{P2}
\bl
\label{sq:p2}
For all $\act_1, \act_2, \actb_1, \actb_2 \in \Act$ and $X \sbset \Stt$,
\bit
\bl
$\lotto{\act_1}{X}{\actb_1} \leqact \lotto{\act_2}{X}{\actb_1}$
 iff  \\
$\lotto{\act_1}{X}{\actb_2} \leqact \lotto{\act_2}{X}{\actb_2}$. 
\eit

\renewcommand{\theenumi}{P3}
\bl
\label{sq:p3}
For all $X \sbset \Stt$, 
if 
there exist some $\act_1, \act_2 \in \Act$ such that 
for all $\actb \in \Act$, 
$\lotto{\act_1}{X}{\actb} \lsact \lotto{\act_2}{X}{\actb}$,
then
for all $\csq_1, \csq_2 \in \Csq$, 
\commentout{
\bit
\bl
$\csq_1 \leqact \csq_2$ iff 
for all $\act \in \Act$, \\
$\lotto{\csq_1}{X}{\act} \leqact \lotto{\csq_2}{X}{\act}$.
\eit
}%
$\csq_1 \leqact \csq_2$ iff 
\bit
\bl
for all $\act \in \Act$, 
$\lotto{\csq_1}{X}{\act} \lsact \lotto{\csq_2}{X}{\act}$, or
\bl
for all $\act \in \Act$, 
$\lotto{\csq_1}{X}{\act} \eeq_\Act \lotto{\csq_2}{X}{\act}$.
\eit

\renewcommand{\theenumi}{P4}
\bl
\label{sq:p4}
For all $X_1, X_2 \sbset \Stt$, $\csq_1, \csqd_1, \csq_2, \csqd_2 \in \Csq$, \\
if
$\csqd_1 \lsact \csq_1$ and $\csqd_2 \lsact \csq_2$, then
\bit
\bl
$\lotto{\csq_1}{X_1}{\csqd_1} \leqact \lotto{\csq_1}{X_2}{\csqd_1}$ iff \\
$\lotto{\csq_2}{X_1}{\csqd_2} \leqact \lotto{\csq_2}{X_2}{\csqd_2}$.
\eit

\renewcommand{\theenumi}{P5}
\bl
\label{sq:p5}
There exist $\csq_1, \csq_2 \in \Csq$ such that $\csq_1 \lsact \csq_2$.

\renewcommand{\theenumi}{P6}
\bl
\label{sq:p6}
For all $\act, \actb \in \Act$, 
$\csq \in \Csq$,
if $\act \lsact \actb$, then 
there exists a
partition $Z_1, \ldots, Z_n$ of $\Stt$, 
such that for all $Z_i$, 
$\lotto{\csq}{Z_{i}}{\act} \lsact \actb$
and 
$\act \lsact \lotto{\csq}{Z_{i}}{\actb}$.

\een

P1 is the standard necessary condition for \representation\ by 
\seu\  (and many of its generalizations), 
since the reals are 
linearly 
ordered;  it basically says that $\leqact$ is a
total preorder.  
More specifically, 
\ref{p1a}
says that $\leqact$ is total (from which
reflexivity follows) and 
\ref{p1b}
says that it is transitive.
P2 
is the \emph{sure-thing principle}, which
says that the way two \acts\ are related depends only on
their differences (that is, the part on which they agree can be ignored, since
that is the ``sure thing'').  
Savage defines
for each subset $X \subseteq S$
a conditional \prefrel\
as follows:  $\act_1 \eleq_\Act^X \act_2$ iff
\bl
for all $\act \in \Act$,
$\lotto{\act_1}{X}{\act} \lsact \lotto{\act_2}{X}{\act}$, or
\bl
for all $\act \in \Act$,
$\lotto{\act_1}{X}{\act} \eeq_\Act \lotto{\act_2}{X}{\act}$.
\eit
Intuitively, $\act_1 \eleq_\Act^X \act_2$ iff when $X$
occurs,
the DM would find $\act_2$ at least as good as $\act_1$.
In the presence of P1,  P2 holds iff
$\eleq_\Act^X$ is a total \preorder\ for all $X$.
Using $\eleq_\Act^X$, Savage defines what it means for $X$ to be null:
$X$ is \emph{null} iff $\act_1 \eleq_\Act^X \act_2$ for all $\act_1,\act_2 \in \Act$\@.
P3 says that if $X$ is not null, then $\eleq_\Act^X$ and $\eleq_\Act$ agree on
the consequences.
Savage defines a relation $\eleq_\Stt$ on events
as follows:  
$X \eleq_\Stt Y$ iff 
for all $\csq,\csqd\in\Csq$,
if $\csqd \lsact \csq$ then
$\lotto{\csq}{X}{\csqd} \leqact \lotto{\csq}{Y}{\csqd}$.
Intuitively, we expect the \dmaker\ to prefer a binary act that is more likely
(according to her beliefs) to yield the more desirable consequence.
} %
The reader might wonder why we do not simply allow $\Act$ to be the set of all
simple acts, since we do not require $\leqact$ to be total.  The point
is that
having $\Act$ consist of all simple acts conceptually requires that the
\dmaker\ explicitly decide, for each pair of acts, whether they are
related, and if so, how; if $\Act$ is a subset of the set of all acts,
then the \dmaker\ does not have to express a preference 
between acts not in $\Act$. 

It turns out that the statement of a number of our results is simpler if
$\Act$ consists of all simple acts.
To  facilitate the comparison of our results with the
standard results from the literature,
where it is typically assumed that $\Act$ consists of all simple acts,
we use 
brackets (\ie\ ``{\ldelim}'' and ``{\rdelim}'')
to delimit parts of the postulates that pertain to 
the general case in which $\Act$ is an arbitrary nonempty subset of the set of
all simple acts.  So there are two versions of the postulates, one for the
general case, which we refer to as the \emph{general version}, and one for the
special case (\ie\ the case in which $\Act$ is the set of all simple acts),
which we refer to as the \emph{special version}.  
The general version includes the bracketed statements while the special version
does not. 
Typically, the statements inside the brackets 
turn unconditional assertions of the special version into implications whose
antecedent says  
that the acts in question are in fact members of $\Act$.
We recommend that the reader ignore the material inside the brackets on
a first pass.
Savage's first six postulates are given in \mfig{fig:savage-postulate}.
It is easy to check that all the bracketed statements are trivially
true if $\Act$ is the set of all simple acts. 
As is standard in the literature,
we use ``$\act_1 \lsact \act_2$'' to abbreviate 
``$\act_1 \leqact \act_2$ and $\act_2 \not\leqact \act_2$'', 
and we use ``$\act_1 \eeq_\Act \act_2$'' to abbreviate
``$\act_1 \leqact \act_2$ and $\act_2 \leqact \act_1$''.
(Note that in general $\lsact$ and $\eeq_\Act$ are not necessarily transitive,
since $\leqact$ is not necessarily transitive.)
Recall that 
$X_1, \ldots, X_n$ 
is a \emph{partition} of $Y$ iff
$\bigcup_i X_i = Y$ and 
for all $1 \leq i,j \leq n$ such
that $i \neq j$, 
$X_i \neq \eset$ and $X_i \cap X_j = \eset$.

\commentout{
\hfill \cond{Mij}.

\hfill [Mij].

(\cond{Mij}

([Mij]

\cond{Test.} Test

[Test.] Test

Test. Test

\hfill \cond{Test} Test

\hfill Test. Test

\hfill \cond{Test.} Test

\hfill \cond{Test.}. Test

\hfill \cond{Test.}\ \ Test

\hfill [Test.] Test

\hfill Test. Test
}%

\bfig
\ben
\renewcommand{\theenumi}{P1}
\bl
\label{p1}
For all $\act_1, \act_2, \act_3 \in \Act$, 
\ben
\bl
\label{p1a}
$\act_1 \leqact \act_2$ or 
$\act_2 \leqact \act_1$, and
\bl
\label{p1b}
if $\act_1 \leqact \act_2$ and $\act_2 \leqact \act_3$, then 
$\act_1 \leqact \act_3$.
\een
\renewcommand{\theenumi}{P2}
\bl
\label{p2}
\commentout{
For all $\act_1, \act_2, \actb_1, \actb_2 \in \Act$ and $X \sbset \Stt$,
if $\lotto{\act_i}{X}{\actb_j} \in \Act$ for $i,j \in \{1, 2\}$, then 
\bit
\bl[]
$\lotto{\act_1}{X}{\actb_1} \leqact \lotto{\act_2}{X}{\actb_1}$ iff 
$\lotto{\act_1}{X}{\actb_2} \leqact \lotto{\act_2}{X}{\actb_2}$. 
\eit 
}%
For all 
$X \sbset \Stt$,
$\act_1, \act_2, \actb_1, \actb_2 \in \Act$, \cond{if 
$\lotto{\act_i}{X}{\actb_j} \in \Act$ for $i,j \in \{1, 2\}$, then}
\bit
\bl[]
$\lotto{\act_1}{X}{\actb_1} \leqact \lotto{\act_2}{X}{\actb_1}$ iff
$\lotto{\act_1}{X}{\actb_2} \leqact \lotto{\act_2}{X}{\actb_2}$.
\eit
\commentout{
For all 
$\lotto{\act_1}{X}{\actb_1}, \lotto{\act_2}{X}{\actb_1},
\lotto{\act_1}{X}{\actb_2}, \lotto{\act_2}{X}{\actb_2} \in \Act$, 
$$\mbox{
$\lotto{\act_1}{X}{\actb_1} \leqact \lotto{\act_2}{X}{\actb_1}$ iff
$\lotto{\act_1}{X}{\actb_2} \leqact \lotto{\act_2}{X}{\actb_2}$.}$$
}%

\renewcommand{\theenumi}{P3}
\bl
\label{p3}
\commentout{
For all $X \sbset \Stt$, 
if 
there exist some $\act_1, \act_2 \in \Act$ such that 
for all $\actb \in \Act$,
$\lotto{\act_1}{X}{\actb} \lsact \lotto{\act_2}{X}{\actb}$,
then
for all $\csq_1, \csq_2 \in \Csq$, 
\bit
\bl[]
$\csq_1 \leqact \csq_2$ iff 
for all $\act \in \Act$, 
$\lotto{\csq_1}{X}{\act} \leqact \lotto{\csq_2}{X}{\act}$.
\eit
}%
For all $X \sbset \Stt$, 
if 
there exist $\act_1, \act_2 \in \Act$ such that 
\cond{there exists $\actb_0 \in \Act$ such that
$\lotto{\act_i}{X}{\actb_0} \in \Act$ 
for $i \in \{1,2\}$, and}
\bit
\bl[]
for all $\actb \in \Act$,
\cond{if
$\lotto{\act_i}{X}{\actb} \in \Act$
for $i \in \{1, 2\}$,  then}
$\lotto{\act_1}{X}{\actb} \lsact \lotto{\act_2}{X}{\actb}$,
\eit
then
for all $\csq_1, \csq_2 \in \Csq$,
\cond{if $\csq_1, \csq_2 \in \Act$, then}
\commentout{
$\csq_1 \lsact \csq_2$ iff 
\cond{there exists $\actb_0 \in \Act$ such that
$\lotto{\csq_i}{X}{\actb_0} \in \Act$
for $i \in \{1, 2\}$, and}
\bit
\bl[]
for all $\actb \in \Act$, 
\cond{if $\lotto{\csq_i}{X}{\actb} \in \Act$ for $i \in \{1, 2\}$, then}
$\lotto{\csq_1}{X}{\actb} \lsact \lotto{\csq_2}{X}{\actb}$ 
\eit
}%
$\csq_1 \leqact \csq_2$ iff 
\cond{there exists $\actb_0 \in \Act$ such that
$\lotto{\csq_i}{X}{\actb_0} \in \Act$
for $i \in \{1, 2\}$, and}
\commentout{
\bit
\bl[]
for all $\actb \in \Act$, 
\cond{if $\lotto{\csq_i}{X}{\actb} \in \Act$ for $i \in \{1, 2\}$, then}
$\lotto{\csq_1}{X}{\actb} \lsact \lotto{\csq_2}{X}{\actb}$ or
\bl[]
for all $\actb \in \Act$, 
\cond{if $\lotto{\csq_i}{X}{\actb} \in \Act$ for $i \in \{1, 2\}$, then}
$\lotto{\csq_1}{X}{\actb} \eeq_\Act \lotto{\csq_2}{X}{\actb}$.
}%
\ben
\bl[]
for all $\actb \in \Act$, 
\cond{if $\lotto{\csq_i}{X}{\actb} \in \Act$ for $i \in \{1, 2\}$, then}
$\lotto{\csq_1}{X}{\actb} \leqact \lotto{\csq_2}{X}{\actb}$.
\een

\commentout{
For all $X \sbset \Stt$, if 
there exist some $\act_1, \act_2, \actb \in \Act$ such that 
$\lotto{\act_1}{X}{\actb} \lsact \lotto{\act_2}{X}{\actb}$,
then 
\bit
\bl
for all $\csq_1, \csq_2 \in \Csq$, 
$\csq_1 \leqact \csq_2$ iff for all 
$\lotto{\csq_1}{X}{\act}, \lotto{\csq_2}{X}{\act} \in \Act$,
$\lotto{\csq_1}{X}{\act} \leqact \lotto{\csq_2}{X}{\act}$.
\eit
}%

\renewcommand{\theenumi}{P4}
\bl
\label{p4}
\commentout{
For all $X_1, X_2 \sbset \Stt$, $\csq_1, \csqd_1, \csq_2, \csqd_2 \in \Csq$, 
if
$\csqd_1 \lsact \csq_1$, and $\csqd_2 \lsact \csq_2$, then
\bit
\bl[]
$\lotto{\csq_1}{X_1}{\csqd_1} \leqact \lotto{\csq_1}{X_2}{\csqd_1}$ iff 
$\lotto{\csq_2}{X_1}{\csqd_2} \leqact \lotto{\csq_2}{X_2}{\csqd_2}$.
\eit
}%
For all $X_1, X_2 \sbset \Stt$, $\csq_1, \csqd_1, \csq_2, \csqd_2 \in \Csq$, 
if \cond{$\csq_1, \csqd_1, \csq_2, \csqd_2 \in \Act$,}
$\csqd_1 \lsact \csq_1$ and $\csqd_2 \lsact \csq_2$, then 
\commentout{
$\lotto{\csq_1}{X_1}{\csqd_1}, \lotto{\csq_1}{X_2}{\csqd_1}, \lotto{\csq_2}{X_1}{\csqd_2} \lotto{\csq_2}{X_2}{\csqd_2} \in \Act$, then 
}%
\cond{if 
$\lotto{\csq_i}{X_j}{\csqd_i} \in \Act$ for $i,j \in \{1, 2\}$, then}
\bit
\bl[]
$\lotto{\csq_1}{X_1}{\csqd_1} \leqact \lotto{\csq_1}{X_2}{\csqd_1}$ iff
$\lotto{\csq_2}{X_1}{\csqd_2} \leqact \lotto{\csq_2}{X_2}{\csqd_2}$.
\eit

\renewcommand{\theenumi}{P5}
\bl
\label{p5}
There exist $\csq_1, \csq_2 \in \Csq$ such that 
\cond{$\csq_1, \csq_2 \in \Act$ and}
$\csq_1 \lsact \csq_2$.

\renewcommand{\theenumi}{P6}
\bl
\label{p6}
\commentout{
For all $\act, \actb \in \Act$, 
$\csq \in \Csq$,
if $\act \lsact \actb$, then 
there exists a
partition $Z_1, \ldots, Z_n$ of $\Stt$ 
all of whose elements are nonempty
such that for all $Z_i$, 
$\lotto{\csq}{Z_{i}}{\act} \lsact \actb$
and 
$\act \lsact \lotto{\csq}{Z_{i}}{\actb}$.
}%
For all $\act, \actb \in \Act$, 
$\csq \in \Csq$,
if $\act \lsact \actb$, then 
there exists a
partition $Z_1, \ldots, Z_n$ of $\Stt$,
such that for all $Z_i$, 
\bit
\bl[]
\cond{if $\lotto{\csq}{Z_{i}}{\act} \in \Act$ then}
$\lotto{\csq}{Z_{i}}{\act} \lsact \actb$
and  
\bl[]
\cond{if $\lotto{\csq}{Z_{i}}{\actb} \in \Act$ then}
$\act \lsact \lotto{\csq}{Z_{i}}{\actb}$.
\eit
\commentout{
For all $\act_1, \act_2 \in \Act$, $\csq \in \Csq$,
if $\act_1 \lsact \act_2$, then there exists a
partition $X_1, \ldots, X_n$ of $\Stt$ such that
if 
$Y_{1,1}, \ldots, Y_{1,k_1}, \ldots, Y_{n,1}, \ldots, Y_{n,k_n}$
is a refinement of the partition, then for all $i, j$,
if $\lotto{\csq}{Y_{i,j}}{\act_1} \notincomp_\Act \act_2$ then
$\lotto{\csq}{Y_{i,j}}{\act_1} \lsact \act_2$ 
and if
$\act_1 \notincomp_\Act \lotto{\csq}{Y_{i,j}}{\act_2}$ then
$\act_1 \lsact \lotto{\csq}{Y_{i,j}}{\act_2}$.
}%

\commentout{
\renewcommand{\theenumi}{P7}
\bl
\label{p7}
For all $\act_1, \act_2 \in \Act$,  $X \sbset \Stt$, if 
for all $\stt \in X$, $\act \in \Act$, 
$\lotto{\act_1}{X}{\act}$ $\leqact (\geqact)$ $\lotto{\act_2(\stt)}{X}{\act}$,
then 
for all $\act \in \Act$,
$\lotto{\act_1}{X}{\act}$ $\leqact  (\geqact)$ $\lotto{\act_2}{X}{\act}$.
}%
\een

\hrule

\caption{Savage's Postulates}
\label{fig:savage-postulate}

\efig

We now give a brief overview of the intuition behind the postulates and how
Savage uses them.
P1 is the standard necessary condition for \representation\ by 
\seu\ (and many of its generalizations), 
since $\realNum$ is a linear order;  it basically says that $\leqact$ is a
total preorder.
Savage defines
for each subset $X \subseteq S$
a 
conditional \prefrel\ 
on acts
as follows:
$\act_1 \eleq_\Act^X \act_2$ iff
\cond{there exists $\act \in \Act$ such that 
$\lotto{\act_i}{X}{\act} \in \Act$ for $i \in \{1, 2\}$ and}
\commentout{
\bit
\bl[] for all $\act \in \Act$, 
\cond{if 
$\lotto{\act_i}{X}{\act} \in \Act$ for $i \in \{1, 2\}$, then}
$\lotto{\act_1}{X}{\act} \lsact \lotto{\act_2}{X}{\act}$ or
\bl[] for all $\act \in \Act$, 
\cond{if 
$\lotto{\act_i}{X}{\act} \in \Act$ for $i \in \{1, 2\}$, then}
$\lotto{\act_1}{X}{\act} \eeq_\Act \lotto{\act_2}{X}{\act}$.
\eit
}%
\ben
\bl[] for all $\act \in \Act$, 
\cond{if 
$\lotto{\act_i}{X}{\act} \in \Act$ for $i \in \{1, 2\}$, then}
$\lotto{\act_1}{X}{\act} \leqact \lotto{\act_2}{X}{\act}$
\een
\commentout{
Note that it follows from P2 that
either 
\bit
\bl
for all $\act \in \Act$, 
\cond{if 
$\lotto{\act_i}{X}{\act} \in \Act$ for $i \in \{1, 2\}$, then}
$\lotto{\act_2}{X}{\act} \leqact \lotto{\act_1}{X}{\act}$ or
\bl
for all $\act \in \Act$, 
$\lotto{\act_2}{X}{\act} \not\leqact \lotto{\act_1}{X}{\act}$.
\eit
}%
(As in the statements of the postulates, we use brackets to delimit parts that
are needed for the general version.)
Intuitively, $\act_1 \eleq_\Act^X \act_2$ 
if when $X$
occurs
the \dmaker\ would find $\act_2$ at least as good as $\act_1$.
Note that ${\leqact} = {\eleq_\Act^\Stt}$, so P1 guarantees that 
$\eleq_\Act^\Stt$ is a total preorder.  However, 
$\eleq_\Act^X$ is
not necessarily a total preorder for all $X$, even if P1 holds---for this, we
need P2.

P2
says that the way two \acts\ are related depends only on
where they differ; the part on which they agree can be ignored.
\commentout{
P2 basically ensures that 
$\lotto{\act_1}{X}{\act} \leqact \lotto{\act_2}{X}{\act}$ for some $\act \in \Act$, then 
$\lotto{\act_1}{X}{\act} \leqact \lotto{\act_2}{X}{\act}$ for all 
$\act \in \Act$.
}%
\commentout{
P1 together with 
P2 allow us to define,
}%
\commentout{
Savage defines
for each subset $X \subseteq S$
a 
conditional \prefrel\ 
on acts
as follows:
$\act_1 \eleq_\Act^X \act_2$ iff
\cond{there exists $\act \in \Act$ such that 
$\lotto{\act_i}{X}{\act} \in \Act$ for $i \in \{1,2\}$ and}
for all $\act \in \Act$, 
\cond{if 
$\lotto{\act_i}{X}{\act} \in \Act$ for $i \in \{1,2\}$, then}
$\lotto{\act_1}{X}{\act} \lsact \lotto{\act_2}{X}{\act}$.
(As in the statements of the postulates, we use brackets to delimit parts that
are needed for the general version.)
Intuitively, $\act_1 \eleq_\Act^X \act_2$ 
if when $X$
occurs
the \dmaker\ would find $\act_2$ at least as good as $\act_1$.
P1 and P2,
together with the assumption that $\Act$ is the set of all simple acts, 
ensure that
$\eleq_\Act^X$ is a total preorder.
}%
Note that it follows from P2 that
either 
\bit
\bl
for all $\act \in \Act$, 
\cond{if 
$\lotto{\act_i}{X}{\act} \in \Act$ for $i \in \{1,2\}$, then}
$\lotto{\act_2}{X}{\act} \leqact \lotto{\act_1}{X}{\act}$ or
\bl
for all $\act \in \Act$, 
$\lotto{\act_2}{X}{\act} \not\leqact \lotto{\act_1}{X}{\act}$.
\eit
Thus, in the presence of P2, $\act_1 \eleq_\Act^X \act_2$ iff
\bit
\bl
for all $\act \in \Act$, 
\cond{if 
$\lotto{\act_i}{X}{\act} \in \Act$ for $i \in \{1,2\}$, then}
$\lotto{\act_2}{X}{\act} \lsact \lotto{\act_1}{X}{\act}$ or
\bl
for all $\act \in \Act$, 
\cond{if 
$\lotto{\act_i}{X}{\act} \in \Act$ for $i \in \{1,2\}$, then}
$\lotto{\act_2}{X}{\act} \eeq_\Act \lotto{\act_1}{X}{\act}$.
\eit
\commentout{
Note that, in the presence of P1,  P2 holds iff
$\eleq_\Act^X$ is a total \preorder\ for all $X$.
Since we do not assume that $\Act$ is the set of all simple acts
(and we do not assume that P1 and P2 necessarily hold), 
$\eleq_\Act^X$ is not necessarily a total preorder here.
}%
Using $\eleq_\Act^X$, Savage defines what it means for a set to be null:
a set $X$ is \emph{null} iff 
 for all $\act_1, \act_2 \in \Act$, 
$\act_1 \eleq_\Act^X \act_2$ \cond{iff $\act_2 \eleq_\Act^X \act_1$}.  
It is easy to check that if $\eleq_\Act^X$ is a total preorder then the
general version and the special version are equivalent. 
Note that $X$ is not null iff there exist
$\act_1, \act_2 \in \Act$ such that $\act_1 \els_\Act^X \act_2$. 
In other words, if $X$ is not null, then the \dmaker\ has some nontrivial
preference if $X$ 
occurs.
P3 basically says that if $X$ is not null, then 
$\csq_1 \leqact \csq_2$ iff $\csq_1 \eleq_\Act^X \csq_2$.
That is, whenever the \dmaker\ has some nontrivial preference, the preferences
over \consequences\ remain the same as the unconditional ones. 
\commentout{
P3 says that, if the \dmaker\
has any strict preferences at all when an event $X$ 
occurs,
then her
preferences over 
the constant \acts\ (which represents her preferences over \consequences)
remains unchanged.
}%
\commentout{
P1--P4 allow
us to define
}%
Savage defines
a relation $\eleq_\Stt$
on events 
as follows:  
$X \eleq_\Stt Y$ 
iff 
\bit
\bl[] 
for all $\csq, \csqd \in \Csq$, 
if 
\cond{$\csq, \csqd \in \Act$,}
$\csqd \lsact \csq$,  
\cond{and 
$\lotto{\csq}{X}{\csqd}, \lotto{\csq}{Y}{\csqd} \in \Act$,}  then 
$\lotto{\csq}{X}{\csqd} \leqact \lotto{\csq}{Y}{\csqd}$. 
\eit
The intuition is that, given two \consequences\ $\csq$ and $\csqd$ such that
$\csqd \lsact \csq$, the \dmaker\ prefers a binary act that 
is more likely to yield $\csq$ than $\csqd$, according to her beliefs. 
This is very much in the spirit of arguments of
de Finetti~\citeyear{DeFinetti31}\@.
P4, in the presence of 
P1--P3 and the assumption that $\Act$ is the set of all simple acts, 
basically ensures that 
$\eleq_\Stt$ is
a total preorder.
P5 says that $\Stt$ is not null.  That is, the \dmaker\ has some nontrivial
(unconditional) preference. 
\commentout{
P5 requires that the \dmaker\ has some
nontrivial preference.
}%
P1--P5 by themselves do not allow the construction of 
a unique
\seu\ \representation\ 
(even if we assume that $\Act$ is the set of all simple acts).
However, 
with the assumption that $\Act$ is the set of all simple acts,
P1--P5 ensure that
$\eleq_\Stt$ is a qualitative probability. 
In order to obtain a 
unique
 \seu\ \representation\
we need 
P6, which says roughly that 
\commentout{
we can find finite partitions of $\Stt$
into arbitrarily
unlikely events.  
}%
for all pairs of acts $\act, \actb \in \Act$ and consequences $\csq \in \Csq$,
if $\act \lsact \actb$ then we can partition $\Stt$ into events such that
the \dmaker\ does not care if $\csq$ were to happen in any element of the
partition.
Savage also has a seventh postulate, but it is 
relevant only for general (nonsimple) acts.
\commentout{
it follows from P1 and P3 for simple acts. 
}%
Since we consider only simple acts, we omit it here.

It may seem that we should consider stronger versions of some
postulates in the general case.  For example, we might consider a
version of P2 that says that if $\act_1$, $\act_2$, $\actb_1$, and
$\actb_2$ are simple acts (not necessarily in $\Act$) such that
$\lotto{\act_i}{X}{\actb_j} \in \Act$  for $1 \le i, j \le 2$, then 
$\lotto{\act_1}{X}{\actb_1} \leqact \lotto{\act_2}{X}{\actb_1}$
iff 
$\lotto{\act_1}{X}{\actb_2} \leqact \lotto{\act_2}{X}{\actb_2}$. 
Fortunately, it is not hard to show that the stronger version is
equivalent to the version that we have stated here, where $\act_1$,
$\act_2$, $\actb_1$, and $\actb_2$ are required to be in $\Act$, since if
$\lotto{\act}{X}{\actb} \in \Act$,
then in fact there exist some 
$\act', \actb' \in \Act$ such that $\lotto{\act'}{X}{\actb'} = \lotto{\act}{X}{\actb}$:
just take $\act' = \actb' = \lotto{\act}{X}{\actb}$. Thus it suffices in all
the postulates to quantify over $\Act$ instead of over all simple acts.

Given a \dprob\ $\dprb = ((\Act,\Stt,\Csq),\edom,\utf,\plf)$  
and 
$\eset \neq Z \sbset \Stt$, 
define the \gseu\ of act
$\act$ \emph{restricted to} $Z$ 
as follows:
\[
\eplutcond{\act}{Z} = 
\EP_{x \in \utf_{\act}(Z)}
\plf(\utf_{\act}^{-1}(x) \inter Z) \et x,
\]
Note that $\eplutcond{\act}{\Stt} = \eplut{\act}$.  
\commentout{
Since we are restricting to additive decision problems
in this section,
it is easy to check that, 
}%
Suppose that $\dprb = ((\Act,\Stt,\Csq),\edom,\utf,\plf)$ is additive.
It is 
then
easy to check that, 
for all \npsubs\ $X$ of $\Stt$, 
\[
\eplut{\act} = \eplutcond{\act}{\Stt} = \eplutcond{\act}{X} \ep \eplutcond{\act}{\comp{X}},
\]
and, more
generally, 
given 
a partition $X_1, \ldots, X_n$ of $Y \sbset \Stt$,
we have that
\[
\eplutcond{\act}{Y} = \eplutcond{\act}{X_1} \ep \cdots \ep \eplutcond{\act}{X_n}.
\]
Also, it is easy to check that for all \npsubs\ $X$ of $\Stt$,
\[
\eplut{\lotto{\act_1}{X}{\act_2}} = \eplutcond{\act_1}{X} \ep \eplutcond{\act_2}{\comp{X}}.
\]
Note that while these statements are true for additive \dprobs, they are not true in general.

\bfig
\ben
\renewcommand{\theenumi}{\axn{\arabic{enumi}}}
\item 
\label{a1}
\commentout{
For all $\eua_1, \eua_2, \eua_3 \in \Eut$, 
$\ev{\eua_1} \eleq_\eudom \ev{\eua_2}$ or $\ev{\eua_2} \eleq_\eudom
\ev{\eua_1}$, and  
$\ev{\eua_2} \eleq_\eudom \ev{\eua_3}$, then $\ev{\eua_1} \eleq_\eudom
\ev{\eua_3}$.
}%
For all $x, y, z \in \Eut(\Stt)$,
\ben
\bl
\label{a1a}
$x \eleq_\eudom y$ or $y \eleq_\eudom x$, and
\bl
\label{a1b}
if $x \eleq_\eudom y$ and $y \eleq_\eudom z$, then $x \eleq_\eudom z$.
\een

\item 
\label{a2}
\commentout{
For all $\eua_1, \eub_1, \eua_2, \eub_2 \in \Eut$, $X \sbset \Stt$, if 
$\ev{\eua_1 \rst X} \ep \ev{\eub_1 \rst \comp{X}} \eleq_\eudom
\ev{\eua_2 \rst X} \ep \ev{\eub_1 \rst \comp{X}}$, then 
$\ev{\eua_1 \rst X} \ep \ev{\eub_2 \rst \comp{X}} \eleq_\eudom 
\ev{\eua_2 \rst X} \ep \ev{\eub_2 \rst \comp{X}}$.  
}%
\commentout{
For all nonempty proper subsets $X \sbset \Stt$, $x_1, x_2 \in \Eut(X)$,
$y_1, y_2 \in \Eut(\comp{X})$, 
\bit
\bl 
$x_1 \ep y_1 \eleq_\eudom x_2 \ep y_1$ iff
$x_1 \ep y_2 \eleq_\eudom x_2 \ep y_2$.
\eit
}%
For all 
\npsubs\ $X$ of $\Stt$, 
$x_1, x_2 \in \Eut(X)$,
$y_1, y_2 \in \Eut(\comp{X})$, 
\cond{if $x_i \ep y_j \in \Eut(\Stt)$ for $i,j \in \{1, 2\}$, then}
\bit
\bl[]
$x_1 \ep y_1 \eleq_\eudom x_2 \ep y_1$ iff
$x_1 \ep y_2 \eleq_\eudom x_2 \ep y_2$.
\eit

\item 
\label{a3}
\commentout{
For all $X \sbset \Stt$, if there exist
$\eua_1, \eua_2 \in \Eut$ such that for all $\eub \in \Eut$, 
$\ev{\eua_1 \rst X} \ep \ev{\eub \rst \comp{X}} \els_\edom \ev{\eua_2
\rst X} \ep \ev{\eub \rst \comp{X}}$, then 
for all $u_1, u_2 \in \ran(\utf)$, 
$u_1 \eleq_\eudom u_2$ iff for all $\eua \in \Eut$, 
$\plf(X) \et u_1 \ep \ev{\eua \rst \comp{X}} \eleq_\eudom \plf(X) \et u_2
\ep \ev{\eua \rst \comp{X}}$. 
}%
For all 
\npsubs\ $X$ of $\Stt$, 
if there exist
$x_1, x_2 \in \Eut(X)$ such that
\cond{there exists
$y_0 \in \Eut(\comp{X})$ such that
$x_i \ep y_0 \in \Eut(\Stt)$ for $i \in \{1,2\}$, and}
\bit
\bl[]
for all $y \in \Eut(\comp{X})$,
\cond{if $x_i \ep y \in \Eut(\Stt)$ for $i \in \{1,2\}$, then}
$x_1 \ep y \els_\eudom x_2 \ep y$, 
\eit
then for all 
$u_1, u_2 \in \ran(\utf)$, 
\cond{if $u_1, u_2 \in \Eut(\Stt)$, then}
$u_1 \eleq_\eudom u_2$ iff
\cond{there exists $y_0 \in \Eut(\comp{X})$ 
such that
$\plf(X) \et u_i \ep y_0 \in \Eut(\Stt)$ for $i \in \{1,2\}$, and}
\bit
\bl[]
for all $y \in \Eut(\comp{X})$, 
\cond{if $\plf(X) \et u_i \ep y \in \Eut(\Stt)$ for $i \in \{1,2\}$, then} \\
$\plf(X) \et u_1 \ep y \eleq_\eudom \plf(X) \et u_2 \ep y$.
\eit

\item 
\label{a4}
For all 
$X_1, X_2 \sbset \Stt$, 
$u_1, v_1, u_2, v_2 \in \ran(\utf)$, 
if 
\cond{$u_1, v_1, u_2, v_2 \in \Eut(\Stt)$,}
$v_1 \els_\eudom u_1$ and $v_2 \els_\eudom u_2$, 
then 
\cond{if $\ulotto{u_i}{X_j}{v_i} \in \Eut(\Stt)$ for 
$i,j \in \{1, 2\}$, then}
\bit
\bl[]
$\ulotto{u_1}{X_1}{v_1} \eleq_\eudom \ulotto{u_1}{X_2}{v_1}$
iff 
$\ulotto{u_2}{X_1}{v_2} \eleq_\eudom \ulotto{u_2}{X_2}{v_2}$.
\eit

\item 
\label{a5}
There exist $u_1, u_2 \in \ran(\utf)$ such that 
\cond{$u_1, u_2 \in \Eut(\Stt)$ and}
$u_1 \els_\eudom u_2$. 

\commentout{
\utf(\csq_1) \els \utf(\csq_2))
There exists $u_1$ and $u_2$ such that $u_1 \lsb u_2$. 
}%

\item  
\label{a6} 
\commentout{
For all $\eua_1, \eua_2 \in \Eut$, 
if ${\eua_1} \els_\edom {\eua_2}$, then 
for all $u \in \ran(\utf)$,
there exist a partition
$X_1, \ldots, X_n$ of $\Stt$ such that for all $X_i$,
\bit
\bl
$\plf(X_i) \et u \ep \ev{\eua_1 \rst \comp{X_i}} \els_\edom \ev{\eua_2}$
and
\bl  
$\ev{\eua_1} \els_\edom \plf(X_i) \et u \ep \ev{\eua_2 \rst \comp{X_i}}$. 
\eit  
}%
For all $x, y \in \Eut(\Stt)$, 
$u \in \ran(\utf)$, 
if $x \els_\eudom y$, then
for all $\act, \actb \in \Act$, $\csq \in \Csq$, such that $\eplut{\act} = x$,
$\eplut{\actb} = y$, and $\utf(\csq) = u$, 
there exists a partition
$Z_1, \ldots, Z_n$ of $\Stt$, such that $x$ can be expressed as
$x_1 \ep \cdots \ep x_n$ and 
$y$ can be expressed as $y_1 \ep \cdots \ep y_n$, where 
$x_i = \eplutcond{\act}{Z_i}$ and 
$y_i = \eplutcond{\actb}{Z_i}$ for $1 \leq i \leq n$,
and
for all 
$1 \leq k \leq n$, 
\bit
\bl[]
\cond{if 
$\plf(Z_k) \et u \ep \EP_{i \neq k} x_i \in \Eut(\Stt)$ 
then}
$\plf(Z_k) \et u \ep \EP_{i \neq k} x_i \els_\eudom y$
and
\bl[]
\cond{if
$\plf(Z_k) \et u \ep \EP_{i \neq k} y_i \in \Eut(\Stt)$ 
then}
$x \els_\eudom \plf(Z_k) \et u \ep \EP_{i \neq k} y_i$.
\eit
\commentout{
For all $\act, \actb \in \Act$, if $\eplut{\act} \els_\eudom \eplut{\actb}$,
then for all $u \in \ran(\utf)$, 
there exists a partition $Z_1, \ldots, Z_n$ of $\Stt$, such that for all $Z_i$,
\bit
\bl 
if $\plf(Z_i) \et u \ep \eplutcond{\act}{Z_i} \in \Eut(\Stt)$ then 
$\plf(Z_i) \et u \ep \eplutcond{\act}{Z_i} \els_\eudom \eplut{\actb}$, and
\bl
if $\plf(Z_i) \et u \ep \eplutcond{\actb}{Z_i} \in \Eut(\Stt)$ then 
$\eplut{\act} \els_\eudom \plf(Z_i) \et u \ep \eplutcond{\actb}{Z_i}$.
\eit
}%
\een

\hrule

\caption{Axioms about Decision Problems}
\label{fig:gseu-axioms}
\efig

Let $\Eut_\dprb(X) = \{ \eplutcond{\act}{X} \st \act \in \Act \}$.  (We omit
the subscript $\dprb$ if it is clear from context.)
Intuitively, $\Eut_\dprb(X)$ consists of all the expected utility values of
acts in $\Act$ restricted to $X$.  
\commentout{
To simplify the statement of one of the axioms,
let
$\ulotto{u}{X}{v} = \plf(X) \et u \ep \plf(\comp{X}) \et v$
if $\eset \neq X \neq \Stt$, and 
let $\ulotto{u}{\eset}{v} = v$ and $\ulotto{u}{\Stt}{v} = u$.

\ben
\renewcommand{\theenumi}{\axn{\arabic{enumi}}}
\item 
\label{sa:p1}
For all $x, y, z \in \Eut(\Stt)$,
\bit
\item[(a)]
$x \eleq_\eudom y$ or $y \eleq_\eudom x$, and
\item[(b)]
if $x \eleq_\eudom y$ and $y \eleq_\eudom z$, then $x \eleq_\eudom z$.
\eit

\item 
\label{sa:p2}
\commentout{
For all $\eua_1, \eub_1, \eua_2, \eub_2 \in \Eut$, $X \sbset \Stt$, if 
$\ev{\eua_1 \rst X} \ep \ev{\eub_1 \rst \comp{X}} \eleq_\eudom
\ev{\eua_2 \rst X} \ep \ev{\eub_1 \rst \comp{X}}$, then 
$\ev{\eua_1 \rst X} \ep \ev{\eub_2 \rst \comp{X}} \eleq_\eudom 
\ev{\eua_2 \rst X} \ep \ev{\eub_2 \rst \comp{X}}$.  
}%
For all 
$\eset \neq X \neq \Stt$, 
$x_1, x_2 \in \Eut(X)$,
$y_1, y_2 \in \Eut(\comp{X})$, 
\bit
\bl 
$x_1 \ep y_1 \eleq_\eudom x_2 \ep y_1$ iff
$x_1 \ep y_2 \eleq_\eudom x_2 \ep y_2$.
\eit

\item 
\label{sa:p3}
\commentout{
For all $X \sbset \Stt$, if there exist
$\eua_1, \eua_2 \in \Eut$ such that for all $\eub \in \Eut$, 
$\ev{\eua_1 \rst X} \ep \ev{\eub \rst \comp{X}} \els_\edom \ev{\eua_2
\rst X} \ep \ev{\eub \rst \comp{X}}$, then 
for all $u_1, u_2 \in \ran(\utf)$, 
$u_1 \eleq_\eudom u_2$ iff for all $\eua \in \Eut$, 
$\plf(X) \et u_1 \ep \ev{\eua \rst \comp{X}} \eleq_\eudom \plf(X) \et u_2
\ep \ev{\eua \rst \comp{X}}$. 
}%
For all $\eset \neq X \neq \Stt$, if
there exist
$x_1, x_2 \in \Eut(X)$ such that
for all $y \in \Eut(\comp{X})$
$x_1 \ep y \els_\eudom x_2 \ep y$, 
then for all $u_1, u_2 \in \ran(\utf)$, 
\commentout{
\bit
\bl
$u_1 \eleq_\eudom u_2$ iff
for all $y \in \Eut(\comp{X})$, \\
$\plf(X) \et u_1 \ep y \eleq_\eudom \plf(X) \et u_2 \ep y$.
\eit
}%
$u_1 \eleq_\eudom u_2$ iff
\bit
\bl
for all $y \in \Eut(\comp{X})$, \\
$\plf(X) \et u_1 \ep y \els_\eudom \plf(X) \et u_2 \ep y$, or
\bl
for all $y \in \Eut(\comp{X})$, \\
$\plf(X) \et u_1 \ep y \eeq_\eudom \plf(X) \et u_2 \ep y$.
\eit

\item 
\label{sa:p4}
For all $X_1, X_2 \sbset \Stt$, $u_1, v_1, u_2, v_2 \in \ran(\utf)$, \\
if 
$v_1 \els_\eudom u_1$ and $v_2 \els_\eudom u_2$, 
then
\commentout{
\bit
\bl
$\plf(X_1) \et u_1 \ep \plf(\comp{X_1}) \et v_1 \eleq_\eudom \plf(X_2) \et u_1 \ep \plf(\comp{X_2}) \et v_1$ 
iff \\
$\plf(X_1) \et u_2 \ep \plf(\comp{X_1}) \et v_2 \eleq_\eudom \plf(X_2) \et u_2 \ep \plf(\comp{X_2}) \et v_2$.
\eit
}%
\bit
\bl
$\ulotto{u_1}{X_1}{v_1} \eleq_\eudom \ulotto{u_1}{X_2}{v_1}$
iff \\
$\ulotto{u_2}{X_1}{v_2} \eleq_\eudom \ulotto{u_2}{X_2}{v_2}$
\eit

\item 
\label{sa:p5}
For some $u_1, u_2 \in \ran(\utf)$, $u_1 \els_\eudom u_2$. 

\item  
\label{sa:p6} 
\commentout{
For all $x, y \in \Eut(\Stt)$, if $x \els_\eudom y$, then
for all $u \in \ran(\utf)$, 
there exists a partition
$Z_1, \ldots, Z_n$ of $\Stt$
and elements $x_1, \ldots, x_n, y_1, \ldots, y_n$ 
such that 
$x_i, y_i \in \Eut(Z_i)$,
$x = x_1 \ep \cdots \ep x_n$, $y = y_1 \ep \cdots \ep y_n$ 
and, for all $k \in \{1, \ldots, n\}$, 
$(\EP_{i \neq k} x_i) \ep \plf(Z_k) \et u \els_\eudom y$ and
$x \els_\eudom (\EP_{i \neq k} x_i) \ep \plf(Z_k) \et u$.
}%
For all $x, y \in \Eut(\Stt)$, 
$u \in \ran(\utf)$, 
if $x \els_\eudom y$, then
for all $\act, \actb \in \Act$, $\csq \in \Csq$, such that $\eplut{\act} = x$,
$\eplut{\actb} = y$, and $\utf(\csq) = u$, 
there exists a partition
$Z_1, \ldots, Z_n$ of $\Stt$, such that $x$ can be expressed as
$x_1 \ep \cdots \ep x_n$ and 
$y$ can be expressed as $y_1 \ep \cdots \ep y_n$, where 
$x_i = \eplutcond{\act}{Z_i}$ and 
$y_i = \eplutcond{\actb}{Z_i}$ for $1 \leq i \leq n$,
and
for all 
$1 \leq k \leq n$, 
\bit
\bl
$\plf(Z_k) \et u \ep \EP_{i \neq k} x_i \els_\eudom y$,
and
\bl
$x \els_\eudom \plf(Z_k) \et u \ep \EP_{i \neq k} y_i$.
\eit
\een
}%
To simplify the statement of one of the axioms,
let
\[
\ulotto{u}{X}{v} = \left\{
\begin{array}{ll}
u & \mbox{if $X = \Stt$,}\\
v & \mbox{if $X = \eset$,}\\
\plf(X) \et u \ep \plf(\comp{X}) \et v & \mbox{otherwise,}
\end{array}
\right.
\]
where $u, v \in \udom$ and $X \sbset \Stt$. 
Note that 
$\eplut{\lotto{\csq}{X}{\csqd}} = \ulotto{\utf(\csq)}{X}{\utf(\csqd)}$. 
The cases $X = \eset$ and $X = \Stt$ must be treated specially, since we
do not 
assume that 
$\bottom \et u$ is the identity for $\ep$.
As with Savage's postulates, we use brackets to delimit parts needed
for the general version.
See \mfig{fig:gseu-axioms} for a list of the axioms.

A1 says that the expected utility 
values
are linearly preordered;
more specifically, 
\ref{a1a}
says that they are totally preordered and 
\ref{a1b}
says that the 
relation
is transitive.
Note 
that A1 does not say that 
the whole \valstr\ is linearly 
preordered\mbox{: that} would
be a sufficient but not a necessary condition for 
$\mr{\gseu}(\dprb)$ to
\satisfy\ P1.  Since we want necessary
and sufficient conditions for our \representation\ results,
some axioms apply only to expected utility 
values
rather than to arbitrary
elements of the \valstr.
\commentout{
The other axioms similarly apply only to expected utility 
values.
}%
\commentout{
As the following theorem shows, 
every 
subset
of $\{\mr{A}1(a), \mr{A}1(b), \ldots, \mr{A}6\}$ \represents\ the corresponding
subset
of $\{\mr{P}1(a), \mr{P}1(b), \ldots, \mr{P}6\}$.
}%

\cond{There is a
technical assumption that we need for some parts of the general version 
of our result.
In general, it might be the case
that $\eplut{\act} \in \Eut(\Stt)$, but $\act \notin \Act$; this could happen
if, even though $\act \notin \Act$, 
there is some $\actb \in \Act$ such that $\eplut{\act} = \eplut{\actb}$.
(Note that $\utf_\act$ is well defined whether or not $\act \in \Act$, so 
it makes sense to talk about $\eplut{\act}$ even if $\act \notin \Act$.)
We say that $\dprb$ is
\whole\ iff this does not happen; more precisely, 
$\dprb = ((\Act,\Stt,\Csq),\edom,\utf,\plf)$ is \emph{\whole} 
iff for all simple acts $\act \in \Csq^\Stt$, $\eplut{\act} \in \Eut(\Stt)$ implies 
$\act \in \Act$.  
A \dprob\ whose set of acts is the set of all simple acts is \whole,
but that is not a necessary condition for a \dprob\ to be whole. 
In general, a \dprob\ 
$\dprb = ((\Act,\Stt,\Csq),\edom,\utf,\plf)$, where
$\edom = (\udom,\pdom,\eudom,\ep,\et)$
is whole iff, for all $x \in \eudom$, either every act with expected
utility $x$ is in $\Act$, or no act with expected utility $x$
is in $\Act$.} 

To simplify the statement of the theorem, let 
$\dprbs_{\mi{all}}$ be the collection of all \plstic\ \dprobs\ and let
$\dprbs_{\mi{add}}$ be the
collection of additive \dprobs.
Also, let $\dprbs_{0}$ be the collection of
\dprobs\ whose set of acts is the set of all simple acts
\cond{along with all \dprobs\ that are whole}.
Let
\bit
\bl 
$\dprbs_{\mr{1a}} = \dprbs_{\mr{1b}} = \dprbs_5 = \dprbs_{\mi{all}}$, 
$\dprbs_4 = \dprbs_0$, and
\bl 
$\dprbs_2 = \dprbs_3 = \dprbs_6 = \dprbs_{\mi{add}} \cap \dprbs_0$.
\eit

\newcounter{savSecNum}
\newcounter{savThmNum}
\setcounter{savSecNum}{\value{section}}
\setcounter{savThmNum}{\value{THEOREM}}
\commentout{
\thm
\label{thm:rep-savage}
\commentout{
For all $i_1, \ldots, i_k \in \{1(a), 1(b), \ldots, 6\}$, 
$\{\mr{A}{i_1}, \ldots, \mr{A}{i_k}\}$ \represents\ 
$\{\mr{P}{i_1}, \ldots, \mr{P}{i_k}\}$.
}%
For all $I \sbset \{1(a), 1(b), \ldots, 6\}$, 
$\{\mr{A}i \st i \in I\}$ \represents\ $\{\mr{P}i \st i \in I\}$.
\ethm

Theorem~\ref{thm:rep-savage} is a strong representation result.  For
example, if we are interested in capturing all of Savage's postulates
but the requirement that $\leqact$ is totally ordered, and instead are
willing to allow it to be partially ordered (a situation explored by
Lehmann \citeyear{Lehmann96}), we simply need to drop axiom A1(b).
Although we have focused here on Savage's postulates, it is
straightforward to represent many of the other standard postulates
considered in the decision theory literature in much the same way.
}%

\thm
\label{thm:rep-savage}
\commentout{
For all $i_1, \ldots, i_k \in \{\mr{1a}, \mr{1b}, \ldots, \mr{6}\}$, 
$\{\mr{A}{i_1}, \ldots, \mr{A}{i_k}\}$ \represents\ 
$\{\mr{P}{i_1}, \ldots, \mr{P}{i_k}\}$.
}%
For all $i_1, \ldots, i_k \in \{\mr{1a}, \mr{1b}, \ldots, \mr{6}\}$, 
$\{\mr{A}{i_1}, \ldots, \mr{A}{i_k}\}$ \represents\ 
$\{\mr{P}{i_1}, \ldots, \mr{P}{i_k}\}$ with respect to
$\dprbs_{i_1} \cap \cdots \cap \dprbs_{i_k}$.
\ethm
\prf  See the appendix.
\eprf

\commentout{
How do we interpret \mthm{thm:rep-savage}?  One way to interpret the result is
that \gseu\ contains \subrules\ that satisfy any subset of Savage's
postulates.\footnote{We skipped P7, but it is not hard to show that we can
\represent\ P7 as well.}  That is, let $\dm_e$ be any subset of
$\{\mr{A}1, \ldots, \mr{A}6\}$ and let $\dm_r$ be the corresponding subset of 
$\{\mr{P}1, \ldots, \mr{P}6\}$.  Let 
$\dprbs_{\dm_e} = \{ \dprb \st \dprb \in \dprbsrel$ and $\dprb$ \satisfies\
$\dm_e\}$. We see that
$\mr{\gseu}^{\dprbs_{\dm_e}}$ is a \drule\ that \represents\ 
$\{ \leqact \st \leqact$ \satisfies\ $\dm_r\}$. 
In particular, 
$\mr{\gseu}^{\dprbs_{\dm_e}}(\dprb)$ \satisfies\ $\dm_r$ for all
$\dprb$ in its domain.

{F}rom a slightly different perspective, 
\mthm{thm:rep-savage} says that, if we make sure 
$\ep$, $\et$, $\utf$, $\plf$, and $\eleq_\eudom$ \satisfy\ some of our axioms, then 
the \prefrels\ we get from \gseu\ will \satisfy\ the corresponding postulates
of Savage.  Conversely, if $\leqact$ \satisfies\ some of Savage's postulates,
then we can \represent\ $\leqact$ using a \dprob\ $\dprb$ such that $\ep$, $\et$,
$\utf$, $\plf$, and $\eleq_\eudom$ of $\dprb$ \satisfy\ the corresponding axioms.
(Actually, our result is slightly stronger than that:  we also show that if a
$\dprb$ \represents\ $\leqact$, then the $\ep$, $\et$,
$\utf$, $\plf$, and $\eleq_\eudom$ of $\dprb$ must  \satisfy\ the corresponding axioms.)
Of course, we could also take another set of postulates.  We picked
Savage's since his is one of the best-known and his postulates do not mention
real numbers (unlike postulates that are based on standard
lotteries). 
}%
Theorem~\ref{thm:rep-savage} is a strong representation result.  For
example, if we are interested in capturing all of Savage's postulates
but the requirement that $\leqact$ is 
a total preorder, 
and instead are
willing to allow it to be 
a partial preorder
(a situation explored by
Lehmann \citeyear{Lehmann1996}), 
we simply need to drop the axiom 
\ref{a1b}.
Although we have focused here on Savage's postulates, it is
straightforward to represent many of the other standard postulates
considered in the decision theory literature in much the same way.

\section{Conclusion}\label{sec:discussion}
We have introduced \gseu, a notion of generalized \seu, and shown that
\gseu\ can (a) represent all preference relations on 
acts
and (b)
be customized to capture any subset of Savage's postulates.  
As we pointed out in the introduction, these results may be of
particular interest 
to designers of software agents, who may want to deal with more general
representations of tastes and beliefs than real-valued utilities and
probabilities.  If beliefs are represented using a plausibility measures
and tastes are represented by a utility function that is not necessarily
real-valued, the problem for the software designer is reduced to finding
appropriate ways of combining plausibility and utility using $\ep$ and
$\et$,  and finding an appropriate order on the resulting expressions.
The results of this paper suggest that rationality
postulates can be captured by choosing $\ep$ and $\et$ so that they
satisfy certain constraints.  The results of 
\cite{CH03b} show that we lose no generality by using \gseu\ to
represent the decision making process; essentially all decision rules
rules can be (ordinally) represented by \gseu. 
Thus, the framework of 
\expstrs\ 
together with \gseu\ provides a useful level of abstraction in which to
study the general  problem of decision making and rules for decision
making 
and a useful conceptual framework for designing decision rules for
software agents.

\subsection*{Acknowledgments}  We thank Duncan Luce and Peter Wakker
for useful comments on a draft of the paper.

\appendix

\section{Proofs}

\commentout{
\ocor{cor:monotonic}
Every \prefrel\ has a \epmono\ \dpadd\ \gseu\ \representation\
with a $\ep$ identity.
\eocor
}%
\setThm{monoSecNum}{monoThmNum}
\cor
Every \prefrel\ has a \epmono\ \dpadd\ \gseu\ \representation\
with a $\ep$ identity.
\ecor
\unsetThm
\prf
Fix some $\ACT = (\Act, \Stt, \Csq)$ and $\leqact$.
Let $\dprb$ be as defined in the proof of \mthm{thm:anyorder}, except that 
$x \eleq_\eudom y$ iff $x = y$ or there exist $\act, \actb \in \Act$ such that
$\act \leqact \actb$ and 
\ben
\bl
$x = \eplut{\act}$ and $y = \eplut{\actb}$, or
\bl
$x = \eplut{\act} \ep z$ and $y = \eplut{\actb} \ep z$ for some $z \in \eudom$.
\een
Recall that $\eplut{\act} = \act$, 
so without case 2, we are back in the situation described in the proof
of \mthm{thm:anyorder}.
Note that, by construction,
the only way that $\eplut{\act} \ep z$ can be an expected utility value is
if $z \sbset \eplut{\act}$, since
${\ep} = {\cup}$ and proper supersets of 
expected utility values cannot be expected utility values.
Thus, if in case 2 both $x$ and $y$ are expected utility values,
then we must in fact be in case 1; case 2 has an effect only when
$x$ and $y$ are not both expected utility values.  
Thus, case 2 does not affect how pairs of expected utility values are
related, so 
we still have that $\mr{\gseu}(\dprb) = {\leqact}$.

To see that $\ep$ is \epmono\ with respect to this definition of
$\eleq_\eudom$, suppose that 
$x \eleq_\eudom y$.  We need to show that $x \ep z \eleq_\eudom y \ep z$. 
If $x = y$, then $x \ep z = y \ep z$, so the conclusion holds.  
Suppose that $x \neq y$. Then 
there exist some $\act, \actb \in \Act$ such that
$\act \leqact \actb$ and either case 1 or case 2 holds.
It is easy to see that in either case,
$x \ep z \eleq_\eudom y \ep z$ by case 2.
Thus $\dprb$ is a \epmono\ \representation\ of $\leqact$
(and, as we have already observed, $\dprb$ is additive).
\eprf

\commentout{
\othm{thm:anyorder'} Given a decision situation 
$\ACT = (\Act,\Stt,\Csq)$, there 
exists a monotonic, additive \expstr\ $\edom$ and a
plausibility measure $\plf$ on $\Stt$ such that,
for every preference
relation $\leqact$ on $\Act$, there exists a utility function
$\utf_{\leqact}$ on 
$\Csq$ and that 
${\leqact} = \mr{\gseu}(\dprb)$, 
where $\dprb = (\ACT,\edom,\utf_{\leqact},\plf)$.
\eothm
}%
\setThm{aoSecNum}{aoThmNum}
\thm Given a decision situation 
$\ACT = (\Act,\Stt,\Csq)$, there 
exists a monotonic, additive \expstr\ $\edom$ and a
plausibility measure $\plf$ on $\Stt$ such that,
for every preference
relation $\leqact$ on $\Act$, there exists a utility function
$\utf_{\leqact}$ on 
$\Csq$ and that 
${\leqact} = \mr{\gseu}(\dprb)$, 
where $\dprb = (\ACT,\edom,\utf_{\leqact},\plf)$.
\ethm
\unsetThm
\prf 
Let $\Pref(\ACT)$ consist of all 
\prefrels\ on $\Act$. We
now modify the construction in Theorem~\ref{thm:anyorder} as follows:
\begin{enumerate}
\item 
$\udom = (\Csq \times 2^{\Pref(\ACT)}, \eleq_\udom)$, 
where $(\csq,X) \eleq_\udom (\csqd,Y)$ iff
$X = \{\leqact\} = Y$ for some ${\leqact} \in \Pref(\ACT)$ and either
$\csq = \csqd$  or $\act_\csq, \act_\csqd \in \Act$ and 
$\act_\csq \leqact \act_\csqd$. 
\item $\pdom = (2^\Stt, \sbset)$.
\item 
$\eudom = (2^{\Stt \times \Csq} \times 2^{\Pref(\ACT)}, \eleq_\eudom)$, where
$(x,X) \eleq_\eudom (y,Y)$ iff $X = \{\leqact\} = Y$ for some ${\leqact}
\in \Pref(\ACT)$ 
and either $x = y$ or $x, y \in \Act$ and $x \leqact y$.  
\commentout{
\item 
\[
(x,p) \ep (y,p' ) =  
\left \{
\begin{array}{ll}
(x \cup y,p) &\mbox{if $p = p'$}\\
(x \cup y, *) &\mbox{if $p \ne p'$}\\
\end{array}\right.
\]
}%
\item $(x,X) \ep (y,Y) = (x \cup y, X \cup Y)$. 
\item 
$X \et (\csq,Y) = (X \times \{\csq\}, Y)$ for $X \sbset \Stt$, 
$\csq \in \Csq$, and
$Y \sbset \Pref(\ACT)$.
\end{enumerate}
The same arguments as in the proof of Theorem~\ref{thm:anyorder},
this construction gives an additive expectation domain.  We can modify
$\eleq_\eudom$ as in Corollary~\ref{cor:monotonic} to make it monotonic.
With a little more effort, we can further modify it so that there is 
a
$\ep$ identity; we omit details here.

Let $\plf(X) = X$. 
Given a preference relation $\leqact$, define the utility function
$\utf_{\leqact}$
by taking $\utf_{\leqact}(c) = (c, \{\leqact\})$.  Again, the same
arguments as those in Theorem~\ref{thm:anyorder} can be used to show
that 
${\leqact} = \mr{\gseu}(\dprb)$, 
where $\dprb = (\ACT,\edom,\utf_{\leqact},\plf)$. 
\eprf

\commentout{
\othm{thm:rep-savage}
\commentout{
For all $i_1, \ldots, i_k \in \{\mr{1a}, \mr{1b}, \ldots, \mr{6}\}$, 
$\{\mr{A}{i_1}, \ldots, \mr{A}{i_k}\}$ \represents\ 
$\{\mr{P}{i_1}, \ldots, \mr{P}{i_k}\}$.
}%
For all $i_1, \ldots, i_k \in \{\mr{1a}, \mr{1b}, \ldots, \mr{6}\}$, 
$\{\mr{A}{i_1}, \ldots, \mr{A}{i_k}\}$ \represents\ 
$\{\mr{P}{i_1}, \ldots, \mr{P}{i_k}\}$ with respect to
$\dprbs_{i_1} \cap \cdots \cap \dprbs_{i_k}$.
\eothm
}%
\setThm{savSecNum}{savThmNum}
\thm
\commentout{
For all $i_1, \ldots, i_k \in \{\mr{1a}, \mr{1b}, \ldots, \mr{6}\}$, 
$\{\mr{A}{i_1}, \ldots, \mr{A}{i_k}\}$ \represents\ 
$\{\mr{P}{i_1}, \ldots, \mr{P}{i_k}\}$.
}%
For all $i_1, \ldots, i_k \in \{\mr{1a}, \mr{1b}, \ldots, \mr{6}\}$, 
$\{\mr{A}{i_1}, \ldots, \mr{A}{i_k}\}$ \represents\ 
$\{\mr{P}{i_1}, \ldots, \mr{P}{i_k}\}$ with respect to
$\dprbs_{i_1} \cap \cdots \cap \dprbs_{i_k}$.
\ethm
\unsetThm
\prf
We 
first establish the result for singleton sets.  
Let $\dprb = (\ACT, \edom, \utf, \plf)$,
where $\ACT = (\Act, \Stt, \Csq)$,  be an arbitrary \dprob.
\cond{As in the statements of the postulates and axioms, we will use brackets to
delimit the parts of the proof that pertain to the conditional versions.}

\bit
\commentout{
\bl A1 \represents\ P1.

Suppose that $\BB{\dprb}$ \satisfies\ A1; 
we need to show that 
$\mr{\gseu}(\dprb)$ \satisfies\ P1.
Let $\act_1, \act_2, \act_3 \in \Act$.  
Let $x_i = \eplut{\act_i}$; clearly $x_1, x_2, x_3 \in \Eut(\Stt)$. 
Since $\BB{\dprb}$ \satisfies\ A1,
\bit
\bl
$x_1 \eleq_\eudom x_2$ or 
$x_2 \eleq_\eudom x_1$, and
\bl
if $x_1 \eleq_\eudom x_2$ and 
$x_2 \eleq_\eudom x_3$,  then
$x_1 \eleq_\eudom x_3$.  
\eit
In other words,
\bit
\bl
$\act_1 \gleq \act_2$ or 
$\act_2 \gleq \act_1$, and 
\bl
if $\act_1 \gleq \act_2$ and
$\act_2 \gleq \act_3$, then
$\act_1 \gleq \act_3$.
\eit
Thus $\mr{\gseu}(\dprb)$ \satisfies\ P1.

Now suppose that $\mr{\gseu}(\dprb)$ \satisfies\ P1. We need to show that 
$\BB{\dprb}$ \satisfies\ A1.  Let $x_1, x_2, x_3 \in \Eut(\Stt)$.
Then
there exist 
$\act_1, \act_2, \act_3 \in \Act$ such that 
$\eplut{\act_i} = x_i$.
Since $\mr{\gseu}(\dprb)$ \satisfies\ 
P1, 
\bit
\bl
$\act_1 \gleq \act_2$ or $\act_2 \gleq \act_1$, and 
\bl
if $\act_1 \gleq \act_2$ and
$\act_2 \gleq \act_3$, then
$\act_1 \gleq \act_3$.  
\eit
Thus
\bit
\bl
$x_1 \eleq_\eudom x_2$ or 
$x_2 \eleq_\eudom x_1$, and
\bl
if $x_1 \eleq_\eudom x_2$ and 
$x_2 \eleq_\eudom x_3$,  then
$x_1 \eleq_\eudom x_3$. 
\eit
So $\BB{\dprb}$ \satisfies\ A1.
}%
\bl \ref{a1a} \represents\ \ref{p1a} and
with respect to $\dprbs_{\mr{1a}}$
and
\ref{a1b} \represents\ \ref{p1b}
with respect to $\dprbs_{\mr{1b}}$.

We do the case of \ref{a1a} \represents\ \ref{p1a} 
with respect to $\dprbs_{\mr{1a}}$
and leave the
other case, which is completely analogous to the one we do, to the reader.

Suppose that $\BB{\dprb}$ \satisfies\ \ref{a1a}.
We need to show that 
$(\ACT,\mr{\gseu}(\dprb))$ 
\satisfies\ \ref{p1a}.
Let $\act_1, \act_2, \act_3 \in \Act$.  
Let $x_i = \eplut{\act_i}$; clearly $x_1, x_2, x_3 \in \Eut(\Stt)$. 
Since $\BB{\dprb}$ \satisfies\ \ref{a1a},
$x_1 \eleq_\eudom x_2$ or 
$x_2 \eleq_\eudom x_1$.
In other words,
$\act_1 \gleq \act_2$ or 
$\act_2 \gleq \act_1$.
Thus $(\ACT,\mr{\gseu}(\dprb))$ \satisfies\ \ref{p1a}.
Now suppose that 
$(\ACT,\mr{\gseu}(\dprb))$ \satisfies\ \ref{p1a}. We need to show that 
$\BB{\dprb}$ \satisfies\ \ref{a1a}.  Let $x_1, x_2, x_3 \in \Eut(\Stt)$.
Then
there exist 
$\act_1, \act_2, \act_3 \in \Act$ such that 
$\eplut{\act_i} = x_i$.
Since $(\ACT,\mr{\gseu}(\dprb))$ \satisfies\ 
\ref{p1a}, 
$\act_1 \gleq \act_2$ or $\act_2 \gleq \act_1$.
Thus
$x_1 \eleq_\eudom x_2$ or 
$x_2 \eleq_\eudom x_1$.
So $\BB{\dprb}$ \satisfies\ \ref{a1a}.

\bl A2 \represents\ P2
with respect to $\dprbs_2$.

Throughout this part of the proof, we 
assume that $\dprb \in \dprbs_2$; in particular, we assume that $\dprb$ is
additive and we will use this fact without further comment.

Suppose that $\BB{\dprb}$ \satisfies\ A2.
We need to show that $(\ACT,\mr{\gseu}(\dprb))$ \satisfies\ P2.
Suppose that $X \sbset \Stt$ and
$\act_1, \act_2, \actb_1, \actb_2 \in \Act$.
\cond{Suppose further that 
$\lotto{\act_i}{X}{\actb_j} \in \Act$.}
We need to show that
\[
\lotto{\act_1}{X}{\actb_1} \gleq \lotto{\act_2}{X}{\actb_1}
\mbox{ iff }
\lotto{\act_1}{X}{\actb_2} \gleq \lotto{\act_2}{X}{\actb_2}.
\]
If $X = \eset$ or $X = \Stt$, then the above is trivially true.  
So assume that $X$ is
a \npsub\ of $\Stt$.
Let $x_i = \eplutcond{\act_i}{X}$ and $y_j = \eplutcond{\actb_j}{\comp{X}}$.  
Clearly
$x_i \in \Eut(X)$, $y_j \in \Eut(\comp{X})$, and
$\eplut{\lotto{\act_i}{X}{\actb_j}} = x_i \ep y_j$.
\cond{Furthermore, $x_i \ep y_j \in \Eut(\Stt)$, since 
$\lotto{\act_i}{X}{\actb_j} \in \Act$.}
Since $\dprb$ \satisfies\ A2, 
we have that
\[
x_1 \ep y_1 \eleq_\eudom x_2 \ep y_1 \mbox{ iff }
x_1 \ep y_2 \eleq_\eudom x_2 \ep y_2, 
\]
so 
\[
\lotto{\act_1}{X}{\actb_1} \gleq \lotto{\act_2}{X}{\actb_1}
\mbox{ iff }
\lotto{\act_1}{X}{\actb_2} \gleq \lotto{\act_2}{X}{\actb_2}.
\]
Thus $(\ACT,\mr{\gseu}(\dprb))$ \satisfies\ P2.

\commentout{
\cond{For the following part of the proof, we need to
make an extra assumption for technical 
reasons.  In general, it might be the case
that $\eplut{\act} \in \Eut(\Stt)$, but $\act \notin \Act$; this could happen
if, even though $\act \notin \Act$, 
there is some $\actb \in \Act$ such that $\eplut{\act} = \eplut{\actb}$.
(Note that $\utf_\act$ is well defined whether or not $\act \in \Act$, so 
it makes sense to talk about $\eplut{\act}$ even if $\act \notin \Act$.)
We say that $\dprb$ is
\whole\ iff this does not happen; more precisely, 
$\dprb = ((\Act,\Stt,\Csq),\edom,\utf,\plf)$ is \emph{\whole} 
\label{def:wholeness}%
iff for all simple acts $\act \in \Csq^\Stt$, $\eplut{\act} \in \Eut(\Stt)$ implies 
$\act \in \Act$.  We will mention explicitly when we need to assume that $\dprb$
is \whole.
\Wholeness\ is a rather
innocuous assumption, since if $\eplut{\act} \in \Eut(\Stt)$, then
the relation of $\act$ to all members of $\Act$ 
is
already determined, whether 
or not $\act$ is a member of $\Act$ (since $\act$ is indistinguishable from some
member of $\Act$).
Note that the set of all simple acts is \whole, but
$\Act$ does not have to be the set of all simple acts in
order  for $\dprb$ to be \whole.}
}%

Suppose that $(\ACT,\mr{\gseu}(\dprb))$ \satisfies\ P2. We need to show that 
$\BB{\dprb}$ \satisfies\ A2.  
\cond{We assume for this direction that $\dprb$ is \whole.}
Suppose that $X$ is a 
\npsub\ of $\Stt$,
$x_1, x_2 \in \Eut(X)$, and 
$y_1, y_2 \in \Eut(\comp{X})$.
\cond{Suppose further that $x_i \ep y_j \in \Eut(\Stt)$.}
We need to show that 
\[
x_1 \ep y_1 \eleq_\eudom x_2 \ep y_1 \mbox{ iff }
x_1 \ep y_2 \eleq_\eudom x_2 \ep y_2.
\]
Note that 
there exist
$\act_1, \act_2, \actb_1, \actb_2 \in \Act$ such that 
$\eplutcond{\act_i}{X} = x_i$ and
$\eplutcond{\actb_i}{\comp{X}} = y_i$.
Observe that 
$\eplut{\lotto{\act_i}{X}{\actb_j}} = x_i \ep y_j$.
\cond{Since  $x_i \ep y_j \in \Eut(\Stt)$ and
$\dprb$ is \whole, it follows that
$\lotto{\act_i}{X}{\actb_j} \in \Act$.}
Since 
$(\ACT,\mr{\gseu}(\dprb))$ \satisfies\ P2, 
\[
\lotto{\act_1}{X}{\actb_1} \gleq \lotto{\act_2}{X}{\actb_1}
\mbox{ iff }
\lotto{\act_1}{X}{\actb_2} \gleq \lotto{\act_2}{X}{\actb_2},
\]
so
\[
x_1 \ep y_1 \eleq_\eudom x_2 \ep y_1 \mbox{ iff }
x_1 \ep y_2 \eleq_\eudom x_2 \ep y_2.
\]
Thus $\dprb$ \satisfies\ A2.

\bl A3 \represents\ P3
with respect to $\dprbs_3$.

Throughout this part of the proof, we 
assume that $\dprb \in \dprbs_3$; in particular, we assume that $\dprb$ is
additive and we will use this fact without further comment.

For this part, we will prove
a slightly stronger claim that actually has a shorter proof. 
Note that A3 and P3 are both implications.  
We show that 
$\BB{\dprb}$ \satisfies\ the 
antecedent (consequent) of A3
iff 
$(\ACT,\mr{\gseu}(\dprb))$ \satisfies\ the
antecedent (consequent) of P3.
\cond{We assume that $\dprb$ is \whole\ in these arguments.}
This implies that
$\BB{\dprb}$ \satisfies\ A3 iff
$(\ACT,\mr{\gseu}(\dprb))$ \satisfies\ P3.

Note that P3 quantifies over all subsets of $\Stt$ while A3 quantifies over
only \npsubs\ of $\Stt$.  
It is easy to check that
$\eset$ and $\Stt$  \satisfy\ P3.
(More precisely, 
$(\ACT,\mr{\gseu}(\dprb))$ \satisfies\ the instance of P3 in which
$X$ is instantiated with $\eset$ and the instance of P3 in which $X$ is
instantiated with $\Stt$.)
So for the rest of this part, we restrict our attention to
\npsubs\ of $\Stt$. 

We begin by showing that 
$\BB{\dprb}$ \satisfies\ the 
antecedent of A3
iff 
$(\ACT,\mr{\gseu}(\dprb))$ \satisfies\ the
antecedent of P3.  
Fix some $X$ that is a \npsub\ of $\Stt$.
We need to show that
there exist 
$x_1, x_2 \in \Eut(X)$ such that 
\bit
\bl[1.]
\cond{there exists $y_0 \in \Eut(\comp{X})$ such that 
$x_i \ep y_0 \in \Eut(\Stt)$ and}
for all $y \in \Eut(\comp{X})$,
\cond{if $x_i \ep y \in \Eut(\Stt)$, then}
$x_1 \ep y \els_\eudom x_2 \ep y$
\eit
iff
there exist $\act_1, \act_2 \in \Act$ such that 
\bit
\bl[2.]
\cond{there exists $\actb_0 \in \Act$ such that
$\lotto{\act_i}{X}{\actb_0} \in \Act$ 
and}
for all $\actb \in \Act$,
\cond{if
$\lotto{\act_i}{X}{\actb} \in \Act$,  then}
$\lotto{\act_1}{X}{\actb} \lsact \lotto{\act_2}{X}{\actb}$.
\eit

To see that 1 implies 2, suppose that 
$x_1, x_2 \in \Eut(X)$ \satisfy\ 1.
Since $x_1, x_2 \in \Eut(X)$, 
there exist $\act_1, \act_2 \in \Act$ such that 
$\eplutcond{\act_i}{X} = x_i$.  
We show that $\act_1$ and $\act_2$ 
\satisfy\ 2. \cond{To see that the first conjunct is true, 
note that by 1 there exists
$y_0 \in \Eut(\comp{X})$ such that  
$x_i \ep y_0 \in \Eut(\Stt)$; fix some such $y_0$.
Note that there exists $\actb_0 \in \Act$ such
that $\eplutcond{\actb_0}{\comp{X}} = y_0$; observe that
$\eplut{\lotto{\act_i}{X}{\actb_0}} = x_i \ep y_0 \in \Eut(\Stt)$. 
Since
$\dprb$ is \whole, $\lotto{\act_i}{X}{\actb_0} \in \Act$.
For the second conjunct, we proceed as follows.}
Let $\actb \in \Act$
\cond{be such that 
$\lotto{\act_i}{X}{\actb} \in \Act$}.
We need to show that
$\lotto{\act_1}{X}{\actb} \lsact \lotto{\act_2}{X}{\actb}$.
Let $y = \eplutcond{\actb}{\comp{X}}$.  Note that
$y \in \Eut(\comp{X})$ and $\eplut{\lotto{\act_i}{X}{\actb}} = x_i \ep y$.
\cond{Furthermore, since $\lotto{\act_i}{X}{\actb} \in \Act$,
$x_i \ep y \in \Eut(\Stt)$.}
By 1, 
$x_1 \ep y \els_\eudom x_2 \ep y$; thus
$\lotto{\act_1}{X}{\actb} \lsact \lotto{\act_2}{X}{\actb}$.

To see that 2 implies 1, suppose that 
$\act_1, \act_2 \in \Act$ satisfy 2.
Let $x_i = \eplutcond{\act_i}{X}$;
note that $x_i \in \Eut(X)$.
We show that $x_1$ and $x_2$ satisfy 1.  
\cond{To see that the first conjunct is true, note that by 
2 there exists $\actb_0 \in \Act$ such that
$\lotto{\act_i}{X}{\actb_0} \in \Act$; fix some such $\actb_0$. 
Let $y_0 = \eplutcond{\actb_0}{\comp{X}}$.
Note that $x_i \ep y_0 = \eplut{\lotto{\act_i}{X}{\actb}}$,
so
$y_0 \in \Eut(\comp{X})$ and
$x_i \ep y_0 \in \Eut(\Stt)$. 
For the second conjunct, we proceed as follows.}
Let $y \in \Eut(\comp{X})$
\cond{be such that $x_i \ep y \in \Eut(\Stt)$}.
We need to show that
$x_1 \ep y \els_\eudom x_2 \ep y$.  
Note that there exists $\actb \in \Act$ such that
$\eplutcond{\actb}{\comp{X}} = y$; observe that
$\eplut{\lotto{\act_i}{X}{\actb}} = x_i \ep y$.
\cond{Furthermore, since  $x_i \ep y \in \Eut(\Stt)$ and $\dprb$ is \whole,
$\lotto{\act_i}{X}{\actb} \in \Act$.}
By 2,
$\lotto{\act_1}{X}{\actb} \lsact \lotto{\act_2}{X}{\actb}$; thus
$x_1 \ep y \els_\eudom x_2 \ep y$.  

We now show that
$\BB{\dprb}$ \satisfies\ the 
consequent of A3
iff 
$(\ACT,\mr{\gseu}(\dprb))$ \satisfies\ the
consequent of P3.  
We need to show that
\bit
\bl[3.]
for all 
$u_1, u_2 \in \ran(\utf)$, 
\cond{if $u_1, u_2 \in \Eut(\Stt)$, then}
$u_1 \eleq_\eudom u_2$ iff
\cond{there exists $y \in \Eut(\comp{X})$ such that
$\plf(X) \et u_i \ep y \in \Eut(\Stt)$ 
and}
for all $y \in \Eut(\comp{X})$, 
\cond{if $\plf(X) \et u_i \ep y \in \Eut(\Stt)$, then} 
$\plf(X) \et u_1 \ep y \eleq_\eudom \plf(X) \et u_2 \ep y$ 
\eit
iff
\bit
\bl[4.]
for all $\csq_1, \csq_2 \in \Csq$,
\cond{if $\csq_1, \csq_2 \in \Act$, then}
$\csq_1 \gleq \csq_2$ iff 
\cond{there exists $\actb \in \Act$ such that
$\lotto{\csq_i}{X}{\actb} \in \Act$
and}
for all $\actb \in \Act$, 
\cond{if $\lotto{\csq_i}{X}{\actb} \in \Act$, then}
$\lotto{\csq_1}{X}{\actb} \gleq \lotto{\csq_2}{X}{\actb}$.
\eit

Suppose that 3 holds. We need to show that 4 holds. 
Fix some $\csq_1, \csq_2 \in \Csq$
\cond{such that $\csq_1, \csq_2 \in \Act$}. 
Let $u_i = \utf(\csq_i)$.  Then
$u_i \in \ran(\utf)$ \cond{and $u_1, u_2 \in \Eut(\Stt)$}. 
Note that $\csq_1 \gleq \csq_2$ iff $u_1 \eleq_\eudom u_2$.  
\commentout{
By 3, $u_1 \els_\eudom u_2$ iff 
\cond{there exists $y \in \Eut(\comp{X})$ such that
$\plf(X) \et u_i \ep y \in \Eut(\Stt)$ 
and} either 3a or 3b hold.
\cond{It is easy to check that 
there exists $y \in \Eut(\comp{X})$ such that
$\plf(X) \et u_i \ep y \in \Eut(\Stt)$ iff
there exists $\actb \in \Act$ such that
$\lotto{\csq_i}{X}{\actb} \in \Act$; the ``only if'' part depends on the
assumption that $\dprb$ is \whole.}
We show that 3a holds iff 4a holds.
(The claim that 3b holds iff 4b holds is established using an analogous argument
and we leave it to the reader.)
}%
By 3, $u_1 \eleq_\eudom u_2$ iff 
\cond{there exists $y \in \Eut(\comp{X})$ such that
$\plf(X) \et u_i \ep y \in \Eut(\Stt)$ 
and} 
for all $y \in \Eut(\comp{X})$, 
\cond{if $\plf(X) \et u_i \ep y \in \Eut(\Stt)$, then} 
$\plf(X) \et u_1 \ep y \eleq_\eudom \plf(X) \et u_2 \ep y$.
\cond{It is easy to check that 
there exists $y \in \Eut(\comp{X})$ such that
$\plf(X) \et u_i \ep y \in \Eut(\Stt)$ iff
there exists $\actb \in \Act$ such that
$\lotto{\csq_i}{X}{\actb} \in \Act$; the ``only if'' part depends on the
assumption that $\dprb$ is \whole.}
To see that 4 holds, 
 fix some $\actb \in \Act$ 
\cond{such that $\lotto{\csq_i}{X}{\actb} \in \Act$}. 
We need to show that $\lotto{\csq_1}{X}{\actb} \gleq \lotto{\csq_2}{X}{\actb}$.
Let $y = \eplutcond{\actb}{\comp{X}}$; then
$y \in \Eut(\comp{X})$ 
and $\plf(X) \et u_i \ep y = \eplut{\lotto{\csq_i}{X}{\actb}}$.
\cond{Since $\lotto{\csq_i}{X}{\actb} \in \Act$, 
 $\plf(X) \et u_i \ep y \in \Eut(\Stt)$}. 
By 3, $\plf(X) \et u_1 \ep y \eleq_\eudom \plf(X) \et u_2 \ep y$;
thus $\lotto{\csq_1}{X}{\actb} \gleq \lotto{\csq_2}{X}{\actb}$.

\commentout{
Suppose that 3a holds.  Fix some $\actb \in \Act$ 
\cond{such that $\lotto{\csq_i}{X}{\actb} \in \Act$}. 
We need to show that $\lotto{\csq_1}{X}{\actb} \gls \lotto{\csq_2}{X}{\actb}$.
Let $y = \eplutcond{\actb}{\comp{X}}$; then
$y \in \Eut(\comp{X})$ 
and $\plf(X) \et u_i \ep y = \eplut{\lotto{\csq_i}{X}{\actb}}$.
\cond{Since $\lotto{\csq_i}{X}{\actb} \in \Act$, 
 $\plf(X) \et u_i \ep y \in \Eut(\Stt)$}. 
By 3a, $\plf(X) \et u_1 \ep y \els_\eudom \plf(X) \et u_2 \ep y$;
thus $\lotto{\csq_1}{X}{\actb} \gls \lotto{\csq_2}{X}{\actb}$.
Suppose that 4a holds. Fix some $y \in \Eut(\comp{X})$
\cond{such that $\plf(X) \et u_i \ep y \in \Eut(\Stt)$}.
We need to show that $\plf(X) \et u_1 \ep y \els_\eudom \plf(X) \et u_2 \ep y$.
Note that there exists some $\actb \in \Act$ such that
$\eplutcond{\actb}{\comp{X}} = y$; observe that
$\eplut{\lotto{\csq_i}{X}{\actb}} = \plf(X)\et u_i \ep y$
\cond{and $\lotto{\csq_i}{X}{\actb} \in \Act$, since 
$\plf(X) \et u_i \ep y \in \Eut(\Stt)$ and
$\dprb$ is \whole}.
By 4a, $\lotto{\csq_1}{X}{\actb} \gls \lotto{\csq_2}{X}{\actb}$;
thus $\plf(X) \et u_1 \ep y \els_\eudom \plf(X) \et u_2 \ep y$.
}%

Now suppose that 4 holds. We need to show that 3 holds. 
Fix some $u_1, u_2 \in \ran(\utf)$
\cond{such that $u_1, u_2 \in \Eut(\Stt)$}. 
Then there exist some $\csq_1, \csq_2 \in \Csq$ such that
$\utf(\csq_i) = u_i$ \cond{and $\csq_1, \csq_2 \in \Act$}. 
Note that $u_1 \eleq_\eudom u_2$ iff
$\csq_1 \gleq \csq_2$. 
By 4, $\csq_1 \gleq \csq_2$ iff
\cond{there exists $\actb \in \Act$ such that
$\lotto{\csq_i}{X}{\actb} \in \Act$
and} 
for all $\actb \in \Act$, 
\cond{if $\lotto{\csq_i}{X}{\actb} \in \Act$, then}
$\lotto{\csq_1}{X}{\actb} \gleq \lotto{\csq_2}{X}{\actb}$.
\cond{As before, it is easy to check that 
there exists $\actb \in \Act$ such that
$\lotto{\csq_i}{X}{\actb} \in \Act$
iff
there exists $y \in \Eut(\comp{X})$ such that
$\plf(X) \et u_i \ep y \in \Eut(\Stt)$;
now the ``if'' part depends on the
assumption that $\dprb$ is \whole.}
To see that 3 holds, fix some $y \in \Eut(\comp{X})$
\cond{such that  $\plf(X) \et u_i \ep y \in \Eut(\Stt)$}.
We need to show that $\plf(X) \et u_1 \ep y \eleq_\eudom \plf(X) \et u_2 \ep y$.
Note that there exists some $\actb \in \Act$ such that
$\eplutcond{\actb}{\comp{X}} = y$; observe that
$\eplut{\lotto{\csq_i}{X}{\actb}} = \plf(X)\et u_i \ep y$
\cond{and $\lotto{\csq_i}{X}{\actb} \in \Act$, since 
$\plf(X) \et u_i \ep y \in \Eut(\Stt)$ and
$\dprb$ is \whole}.
By 
4, $\lotto{\csq_1}{X}{\actb} \gleq \lotto{\csq_2}{X}{\actb}$;
thus $\plf(X) \et u_1 \ep y \eleq_\eudom \plf(X) \et u_2 \ep y$.
\commentout{
Suppose that 3b holds.  Fix some $\actb \in \Act$ 
\cond{such that $\lotto{\csq_i}{X}{\actb} \in \Act$}. 
We need to show that $\lotto{\csq_1}{X}{\actb} \geeq \lotto{\csq_2}{X}{\actb}$.
Let $y = \eplutcond{\actb}{\comp{X}}$; then
$y \in \Eut(\comp{X})$ 
and $\plf(X) \et u_i \ep y = \eplut{\lotto{\csq_i}{X}{\actb}}$.
\cond{Since $\lotto{\csq_i}{X}{\actb} \in \Act$, 
 $\plf(X) \et u_i \ep y \in \Eut(\Stt)$}. 
By 3b, $\plf(X) \et u_1 \ep y \eeq_\eudom \plf(X) \et u_2 \ep y$;
thus $\lotto{\csq_1}{X}{\actb} \geeq \lotto{\csq_2}{X}{\actb}$.
}%

\commentout{
For 
6, 
let $y \in \Eut(\comp{X})$ such that $x_i \ep y \in \Eut(\Stt)$.  
Then there exists some $\actb \in \Act$ such that
$\eplutcond{\actb}{\comp{X}} = y$ and
$x_i \ep y \in \Eut(\Stt)$.  
Since $\eplut{\lotto{\act_i}{X}{\actb}} = x_i \ep y$ and 
$\dprb$ is \whole,
$\lotto{\act_i}{X}{\actb} \in \Act$;
thus $\lotto{\act_1}{X}{\actb} \gls \lotto{\act_2}{X}{\actb}$ by 2, so
$x_1 \ep y \els_\eudom x_2 \ep y$. Since $y$ is arbitrary, 
6 holds.
\commentout{
Note that there exists $\actb \in \Act$ such that 
$\eplutcond{\actb}{\comp{X}} = y$
and $\eplut{\lotto{\act_i}{X}{\actb}} = x_i \ep y$.
Since $\lotto{\act_1}{X}{\actb} \gls \lotto{\act_2}{X}{\actb}$ by assumption,
$x_1 \ep y \els_\eudom x_2 \ep y$.  
Since $y \in \Eut(\comp{X})$ is arbitrary,  we have that, for all $y \in \Eut(\comp{X})$, 
$x_1 \ep y \els_\eudom x_2 \ep y$.
}%

We can apply A3 now and conclude that,
for all 
$u_1, u_2 \in \ran(\utf)$, 
if $u_1, u_2 \in \Eut(\Stt)$, then
$u_1 \eleq_\eudom u_2$ iff
\bit
\bl[7.]
there exists $y \in \Eut(\comp{X})$ such that
$\plf(X) \et u_i \ep y \in \Eut(\Stt)$ for $i \in \{1,2\}$, and
\bl[8.]
for all $y \in \Eut(\comp{X})$, 
if $\plf(X) \et u_i \ep y \in \Eut(\Stt)$ for $i \in \{1,2\}$, then \\
$\plf(X) \et u_1 \ep y \eleq_\eudom \plf(X) \et u_2 \ep y$.
\eit
\commentout{
Now let $\csq_1, \csq_2 \in \Csq$.  Suppose that $\csq_1, \csq_2 \in \Csq$
and let $u_i = \utf(\csq_i)$ for $i = 1, 2$.
For each act $\act \in \Act$,
let $y_{\act} = \eplutcond{\act}{\comp{X}}$.  Note that 
$y_{\act} \in \Eut(\comp{X})$ and each element
in $\Eut(\comp{X})$ has the form $y_{\act}$ for some act $\act \in \Act$.
Thus, it follows from A3 that 
$u_1 \eleq_\eudom u_2$ iff,
for all 
$\act \in \Act$, 
$\plf(X) \et u_1 \ep y_{\act} \eleq_\eudom \plf(X) \et u_2 \ep y_{\act}$.
Thus $\mr{\gseu}(\dprb)$ \satisfies\ P3.
}%
We are ready to show that 3 and 4 hold.  Suppose that 
$\csq_1, \csq_2 \in \Csq$ and $\csq_1, \csq_2 \in \Act$.  Let 
$u_i = \utf(\csq_i)$.  Then $u_1, u_2 \in \ran(\utf)$ and $u_1, u_2 \in \Eut(\Stt)$.
Note that $\csq_1 \gleq \csq_2$ iff
$u_1 \eleq_\eudom u_2$ iff 7 and 8 hold, so it suffices to show that 7 and 8
hold iff 3 and 4 hold.  
\bit
\bl
Suppose that 7 and 8 hold.  Then there exists $y \in \Eut(\comp{X})$ such that
$\plf(X) \et u_i \ep y \in \Eut(\Stt)$. Thus there exists some $\actb \in \Act$
such that $\eplutcond{\actb}{\comp{X}} = y$ and
$\plf(X) \et u_i \ep y \in \Eut(\Stt)$; since $\dprb$ is \whole, 
$\lotto{\csq_i}{X}{\actb} \in \Act$; thus 3 holds.  
For 4, let $\actb \in \Act$ be such that $\lotto{\csq_i}{X}{\actb} \in \Act$.
Let $y = \eplutcond{\actb}{\comp{X}}$.  
Then $y \in \Eut(\comp{X})$ and
$\eplut{\lotto{\csq_i}{X}{\actb}} = \plf(X) \et u_i \ep y \in \Eut(\Stt)$.
By 8, $\plf(X) \et u_1 \ep y \eleq_\eudom \plf(X) \et u_2 \ep y$, so 
$\lotto{\csq_1}{X}{\actb} \gleq \lotto{\csq_2}{X}{\actb}$.  Thus 4 holds.
\bl
Suppose that 3 and 4 hold. Then there exist $\actb \in \Act$ such that
$\lotto{\csq_i}{X}{\actb} \in \Act$.  
Let $y = \eplutcond{\actb}{\comp{X}}$.  
Then $y \in \Eut(\comp{X})$ and
$\eplut{\lotto{\csq_i}{X}{\actb}} = \plf(X) \et u_i \ep y \in \Eut(\Stt)$,
so 7 holds. 
For 8, let $y \in \Eut(\comp{X})$ be such that
 $\plf(X) \et u_i \ep y \in \Eut(\Stt)$.  Then there exists
$\actb \in \Act$ such that $\eplutcond{\actb}{\comp{X}} = y$.
Since $\dprb$ is \whole, $\lotto{\csq_i}{X}{\actb} \in \Act$.  By
4, $\lotto{\csq_1}{X}{\actb} \gleq \lotto{\csq_2}{X}{\actb}$, so
$\plf(X) \et u_1 \ep y \eleq_\eudom \plf(X) \et u_2 \ep y$.  Thus 8 holds.
\eit
\commentout{
Suppose that $\csq_1 \gleq \csq_2$.
Then $u_1 \eleq_\eudom u_2$.  So for all
$y \in \Eut(\comp{X})$, 
$\plf(X) \et u_1 \ep y \eleq_\eudom \plf(X) \et u_2 \ep y$.  
Let $\act \in \Act$.  Note that
$\eplut{\lotto{\csq_i}{X}{\act}} = \plf(X) \et u_i \ep y$,
where $y = \eplutcond{\act}{\comp{X}} \in \Eut(\comp{X})$. 
Thus 
$\lotto{\csq_1}{X}{\act} \gleq \lotto{\csq_2}{X}{\act}$.  Since $\act \in \Act$
is arbitrary, 
\bit
\bl
$\csq_1 \eleq_\eudom \csq_2$ only if 
for all $\act \in \Act$,
$\lotto{\csq_1}{X}{\act} \gleq \lotto{\csq_2}{X}{\act}$.
\eit
Now suppose that 
for all $\act \in \Act$,
$\lotto{\csq_1}{X}{\act} \gleq \lotto{\csq_2}{X}{\act}$.  
Let $y = \Eut(\comp{X})$.  Then 
there exists some $\act \in \Act$ such that
$y = \eplutcond{\act}{\comp{X}}$ 
and $\eplut{\lotto{\csq_i}{X}{\act}} = \plf(X) \et u_i \ep y$.
We see that $\plf(X) \et u_1 \ep y \eleq_\eudom \plf(X) \et u_2 \ep y$.
Since $y \in \Eut(\comp{X})$ is arbitrary, for all
$y \in \Eut(\comp{X})$, 
$\plf(X) \et u_1 \ep y \eleq_\eudom \plf(X) \et u_2 \ep y$.  Thus
$u_1 \eleq_\eudom u_2$, and so $\csq_1 \gleq \csq_2$.  So
\bit
\bl
$\csq_1 \eleq_\eudom \csq_2$ if
for all $\act \in \Act$,
$\lotto{\csq_1}{X}{\act} \gleq \lotto{\csq_2}{X}{\act}$.
\eit
}%
}%

\commentout{
Now suppose that $(\ACT,\mr{\gseu}(\dprb))$ \satisfies\ P3. We need to show that 
$\BB{\dprb}$ \satisfies\ A3.  
Suppose that $X$ is a \npsubs\ of $\Stt$ and there exist
$x_1, x_2 \in \Eut(X)$ such that
\cond{there exists $y \in \Eut(\comp{X})$ such that 
$x_i \ep y \in \Eut(\Stt)$ and} 4 holds. 
We need to show that 
for all 
$u_1, u_2 \in \ran(\utf)$, 
\cond{if $u_1, u_2 \in \Eut(\Stt)$, then}
$u_1 \eleq_\eudom u_2$ iff
\cond{there exists $y \in \Eut(\comp{X})$ such that
$\plf(X) \et u_i \ep y \in \Eut(\Stt)$ and}
5 or 6 hold. 
This time we need to apply P3, so we need to show that the antecedent of P3
holds. That is, we must show that there exist
$\act_1, \act_2 \in \Act$ such that 
\cond{there exists $\actb \in \Act$ such that
$\lotto{\act_i}{X}{\actb} \in \Act$ and} 1 holds. 

Fix some $u_1, u_2 \in \ran(\utf)$ \cond{such that $u_1, u_2 \in \Eut(\Stt)$}.
Then there exist $\csq_1, \csq_2 \in \Csq$ 
such that
\cond{$\csq_1,\csq_2 \in \Act$ and}
$\utf(\csq_i) = u_i$.  
\cond{As before, it is easy to check that 
there exists $y \in \Eut(\comp{X})$ such that
$\plf(X) \et u_i \ep y \in \Eut(\Stt)$ iff
there exists $\actb \in
\Act$ such that $\lotto{\csq_i}{X}{\actb} \in \Act$,
since $\dprb$ is \whole.}

Since 5 holds, there exists $y \in \Eut(\comp{X})$ such that 
$x_i \ep y \in \Eut(\Stt)$.
Thus there exists $\actb \in \Act$ such that 
$\eplutcond{\actb}{\comp{X}} = y$.  Since $\dprb$ is \whole,
$\lotto{\act_i}{X}{\actb} \in \Act$; so 1 holds. 
For 2, suppose that $\actb \in \Act$ and $\lotto{\act_i}{X}{\actb} \in \Act$.
Let $y = \eplutcond{\actb}{\comp{X}}$; then $y \in \Eut(\comp{X})$ and
$x_i \ep y = \eplut{\lotto{\act_i}{X}{\actb}} \in \Eut(\Stt)$, so
$x_1 \ep y \els_\eudom x_2 \ep y$ by 6; thus 
$\lotto{\act_1}{X}{\actb} \gls \lotto{\act_2}{X}{\actb}$, so 2 holds.

Let $u_1, u_2 \in \ran(\utf)$ such that $u_1, u_2 \in \Eut(\Stt)$.  We need to
show that 7 and 8 hold of $u_1$ and $u_2$.  
Let $\csq_1, \csq_2 \in \Csq$ be such that $\utf(\csq_i) = u_i$.  
Since $\dprb$ is \whole, $\csq_1, \csq_2 \in \Act$, so P3 applies.
Thus $u_1 \eleq_\eudom u_2$ iff $\csq_1 \gleq \csq_2$ (by definition)
iff 3 and 4 hold of $\csq_1$ and $\csq_2$ 
(by P3)
iff 7 and 8 hold of $u_1$ and $u_2$ (which we have shown in the previous case).  
Thus $\dprb$ \satisfies\ A3.
\commentout{
Now suppose that $\mr{\gseu}(\dprb)$ \satisfies\ P3. We need to show that 
$\BB{\dprb}$ \satisfies\ A3.  
Let $X$ be a proper nonempty subset of $\Stt$ such that 
there exist some
$x_1, x_2 \in \Eut(X)$ such that 
for all $y \in \Eut(\comp{X})$, 
$x_1 \ep y \els_\eudom x_2 \ep y$.
We need to show that 
for all $u_1, u_2 \in \ran(\utf)$, 
\bit
\bl
$u_1 \eleq_\eudom u_2$ iff
for all $y \in \Eut(\comp{X})$, 
$\plf(X) \et u_1 \ep y \eleq_\eudom \plf(X) \et u_2 \ep y$.
\eit
Let $\act_1, \act_2 \in \Act$ be such that $\eplutcond{\act_i}{X} = x_i$.
Let $\actb \in \Act$.  We want to show that 
$\lotto{\act_1}{X}{\actb} \gls \lotto{\act_2}{X}{\actb}$ (so we can apply P3).
Let $y = \eplutcond{\actb}{\comp{X}}$.  
Clearly $y \in \Eut(\comp{X})$, 
so $x_1 \ep y \els_\eudom x_2 \ep y$ by assumption. Since
$\eplut{\lotto{\act_i}{X}{\actb}} = x_i \ep y$,
we have
$\lotto{\act_1}{X}{\actb} \gls \lotto{\act_2}{X}{\actb}$.  Since
$\actb \in \Act$ is arbitrary, 
for all $\actb \in \Act$, 
$\lotto{\act_1}{X}{\actb} \gls \lotto{\act_2}{X}{\actb}$.  So 
for all $\csq_1, \csq_2 \in \Csq$, 
\bit
\bl
$\csq_1 \gleq \csq_2$ iff 
for all $\act \in \Act$, 
$\lotto{\csq_1}{X}{\act} \gleq \lotto{\csq_2}{X}{\act}$,
\eit
since $(\ACT,\mr{\gseu}(\dprb))$ \satisfies\ P3.
Let $u_1, u_2 \in \ran(\utf)$.  Let $\csq_1, \csq_2 \in \Csq$ be  such that
$\utf(\csq_i) = u_i$.
By P3, $\csq_1 \gleq \csq_2$ iff for all $\act \in \Act$, 
$\lotto{\csq_1}{X}{\act} \gleq \lotto{\csq_2}{X}{\act}$.
\commentout{
So for all $\act \in \Act$,
$\lotto{\csq_1}{X}{\act} \gleq \lotto{\csq_2}{X}{\act}$.
Let $y \in \Eut(\comp{X})$.  
Then there exists some $\act \in \Act$ such that 
$\eplutcond{\act}{X} = y$ and 
$\eplut{\lotto{\csq_i}{X}{\act}} = \plf(X) \et u_i \ep y$.
We see that $\plf(X) \et u_1 \ep y \eleq_\eudom \plf(X) \et u_2 \ep y$.  Since
$y \in \Eut(\comp{X})$ is arbitrary, 
\bit
\bl
$u_1 \eleq_\eudom u_2$ only if
for all $y \in \Eut(\comp{X})$, 
$\plf(X) \et u_1 \ep y \eleq_\eudom \plf(X) \et u_2 \ep y$.
\eit
Now suppose that 
for all $y \in \Eut(\comp{X})$, 
$\plf(X) \et u_1 \ep y \eleq_\eudom \plf(X) \et u_2 \ep y$.
Let $\act \in \Act$.  We see that $\eplutcond{\act}{\comp{X}} = y$
for some $y \in  \Eut(\comp{X})$
and that $\eplut{\lotto{\csq_i}{X}{\act}} = \plf(X) \et u_i \ep y$.
Then we see that $\lotto{\csq_1}{X}{\act} \gleq \lotto{\csq_2}{X}{\act}$.
Since $\act \in \Act$ is arbitrary, 
for all $\act \in \Act$, $\lotto{\csq_1}{X}{\act} \gleq \lotto{\csq_2}{X}{\act}$.
So $\csq_1 \gleq \csq_2$; thus $u_1 \eleq_\eudom u_2$.  We have then
\bit
\bl
$u_1 \eleq_\eudom u_2$ if
for all $y \in \Eut(\comp{X})$, 
$\plf(X) \et u_1 \ep y \eleq_\eudom \plf(X) \et u_2 \ep y$.
\eit
}%
It immediately follows that $u_1 \eleq_\eudom u_2$ iff
for all $y \in \Eut(\comp{X})$, 
$\plf(X) \et u_1 \ep y \eleq_\eudom \plf(X) \et u_2 \ep y$.
So $\dprb$ \satisfies\ A3.
}%
}%

\bl A4 \represents\ P4
with respect to $\dprbs_4$.

Suppose that $\BB{\dprb}$ \satisfies\ A4.
We need to show that $(\ACT,\mr{\gseu}(\dprb))$ \satisfies\ P4.
Suppose that 
$X_1, X_2 \sbset \Stt$, $\csq_1, \csqd_1, \csq_2, \csqd_2 \in \Csq$, 
\cond{$\csq_1, \csqd_1, \csq_2, \csqd_2 \in \Act$, $\lotto{\csq_i}{X_j}{\csqd_i} \in \Act$,}
$\csqd_1 \lsact \csq_1$, and $\csqd_2 \lsact \csq_2$.
We need to show that
\[
\lotto{\csq_1}{X_1}{\csqd_1} \gleq \lotto{\csq_1}{X_2}{\csqd_1}
\mbox{ iff }
\lotto{\csq_2}{X_1}{\csqd_2} \gleq \lotto{\csq_2}{X_2}{\csqd_2}.
\]
Let $u_i = \utf(\csq_i)$ and $v_i = \utf(\csqd_i)$.
Note that
$u_1, v_1, u_2, v_2 \in \ran(\utf)$
\cond{and $u_1,u_2,v_1,v_2 \in \Eut(\Stt)$}.
Also, $v_i \els_\eudom u_i$, since $\csqd_i \gls \csq_i$.
Note that
$\eplut{\lotto{\csq_i}{X_j}{\csqd_i}} = \ulotto{u_i}{X_j}{v_i}$.
\cond{Since $\lotto{\csq_i}{X_j}{\csqd_i} \in \Act$,
$\ulotto{u_i}{X_j}{v_i} \in \Eut(\Stt)$.}
Since $\BB{\dprb}$ \satisfies\ A4,
\[
\ulotto{u_1}{X_1}{v_1} \eleq_\eudom \ulotto{u_1}{X_2}{v_1}
\mbox{ iff }
\ulotto{u_2}{X_1}{v_2} \eleq_\eudom \ulotto{u_2}{X_2}{v_2},
\]
which means that
\[
\lotto{\csq_1}{X_1}{\csqd_1} \gleq \lotto{\csq_1}{X_2}{\csqd_1}
\mbox{ iff }
\lotto{\csq_2}{X_1}{\csqd_2} \gleq \lotto{\csq_2}{X_2}{\csqd_2}.
\]
Thus, $(\ACT,\mr{\gseu}(\dprb))$ \satisfies\ P4.

\commentout{
$X_1, X_2 \sbset \Stt$ and $\csq_1, \csqd_1, \csq_2, \csqd_2 \in \Csq$
such that 
$\csqd_1 \gls \csq_1$ and
$\csqd_2 \gls \csq_2$.
We need to show that 
\[
\lotto{\csq_1}{X_1}{\csqd_1} \gleq \lotto{\csq_1}{X_2}{\csqd_1}
\mbox{ iff }
\lotto{\csq_2}{X_1}{\csqd_2} \gleq \lotto{\csq_2}{X_2}{\csqd_2}.
\]
Let $u_i = \utf(\csq_i)$ and $v_i = \utf(\csqd_i)$.
Note that
$v_i \els_\eudom u_i$, since $\csqd_i \gls \csq_i$.
Furthermore,
$\plf(X_j) \et u_i \ep \plf(\comp{X_j}) \et v_i = \eplut{\lotto{\csq_i}{X_j}{\csqd_i}}$.
Since $\BB{\dprb}$ \satisfies\ A4, this implies that 
\[
\begin{array}{l}
\plf(X_1) \et u_1 \ep \plf(\comp{X_1}) \et v_1 \eleq_\eudom \plf(X_2) \et u_1 \ep \plf(\comp{X_2}) \et v_1
\mbox{ iff} \\
\plf(X_1) \et u_2 \ep \plf(\comp{X_1}) \et v_2 \eleq_\eudom \plf(X_2) \et u_2
\ep \plf(\comp{X_2}) \et v_2,
\end{array}
\]
which means that
\[
\lotto{\csq_1}{X_1}{\csqd_1} \gleq \lotto{\csq_1}{X_2}{\csqd_1}
\mbox{ iff }
\lotto{\csq_2}{X_1}{\csqd_2} \gleq \lotto{\csq_2}{X_2}{\csqd_2}.
\]
Thus, $(\ACT,\mr{\gseu}(\dprb))$ \satisfies\ P4.
}%

Now suppose that $(\ACT,\mr{\gseu}(\dprb))$ \satisfies\ P4.  We need to show that 
$\BB{\dprb}$ \satisfies\ A4.  
\cond{For this direction, we assume that $\dprb$ is \whole.}
Suppose that $X_1, X_2 \sbset \Stt$,
$u_1, v_1, u_2, v_2 \in \ran(\utf)$,
\cond{$u_1, v_1, u_2, v_2 \in \Eut(\Stt)$,
$\ulotto{u_i}{X_j}{v_i} \in \Eut(\Stt)$,}
$v_1 \els_\eudom u_1$, and $v_2 \els_\eudom u_2$.
We need to show that
\[
\ulotto{u_1}{X_1}{v_1} \eleq_\eudom \ulotto{u_1}{X_2}{v_1}
\mbox{ iff }
\ulotto{u_2}{X_1}{v_2} \eleq_\eudom \ulotto{u_2}{X_2}{v_2}.
\]
Let $\csq_1, \csqd_1, \csq_2, \csqd_2 \in \Csq$ be such that 
\cond{$\csq_1, \csqd_1, \csq_2, \csqd_2 \in \Act$,}
$\utf(\csq_i) = u_i$ and $\utf(\csqd_i) = v_i$.
Then we see that
$\csqd_i \gls \csq_i$, since $v_i \els_\eudom u_i$.
Note that $\ulotto{u_i}{X_j}{v_i} = \eplut{\lotto{\csq_i}{X_j}{\csqd_i}}$.
\cond{Since $\ulotto{u_i}{X_j}{v_i} \in \Eut(\Stt)$ and $\dprb$ is \whole,
$\lotto{\csq_i}{X_j}{\csqd_i} \in \Act$.}
Since $(\ACT,\mr{\gseu}(\dprb))$ \satisfies\ P4,
\[
\lotto{\csq_1}{X_1}{\csqd_1} \gleq \lotto{\csq_1}{X_2}{\csqd_1}
\mbox{ iff }
\lotto{\csq_2}{X_1}{\csqd_2} \gleq \lotto{\csq_2}{X_2}{\csqd_2}, 
\]
which
implies that 
\[
\ulotto{u_1}{X_1}{v_1} \eleq_\eudom \ulotto{u_1}{X_2}{v_1}
\mbox{ iff }
\ulotto{u_2}{X_1}{v_2} \eleq_\eudom \ulotto{u_2}{X_2}{v_2}.
\]
So $\dprb$ \satisfies\ A4.

\bl A5 \represents\ P5
with respect to $\dprbs_5$.

$\BB{\dprb}$ \satisfies\ A5 iff there exist $u_1, u_2 \in \ran(\utf)$ such that
\cond{$u_1, u_2 \in \Eut(\Stt)$ and}
$u_1 \els_\eudom u_2$ iff there exist some $\csq_1, \csq_2 \in \Csq$ such that
\cond{$\csq_1, \csq_2 \in \Act$ and}
$\utf(\csq_1) \els_\eudom \utf(\csq_2)$ iff there exist some
$\csq_1, \csq_2 \in \Csq$ such that 
\cond{$\csq_1, \csq_2 \in \Act$ and}
$\csq_1 \gls \csq_2$ iff
$(\ACT,\mr{\gseu}(\dprb))$ \satisfies\ P5.

\bl A6 \represents\ P6
with respect to $\dprbs_6$.

Throughout this part of the proof, we 
assume that $\dprb \in \dprbs_6$; in particular, we assume that $\dprb$ is
additive and we will use this fact without further comment.

Suppose that $\BB{\dprb}$ \satisfies\ A6. We need to show that
$(\ACT,\mr{\gseu}(\dprb))$ \satisfies\ P6. 
Let $\act, \actb \in \Act$ and $\csq \in \Csq$. Suppose that $\act \gls \actb$.
We need to show that there exists a partition $Z_1, \ldots, Z_n$ of $\Stt$,
such that for all $Z_i$, 
\bit
\bl[1.]
\cond{if $\lotto{\csq}{Z_{i}}{\act} \in \Act$ then}
$\lotto{\csq}{Z_{i}}{\act} \gls \actb$
and  
\bl[2.]
\cond{if $\lotto{\csq}{Z_{i}}{\actb} \in \Act$ then}
$\act \gls \lotto{\csq}{Z_{i}}{\actb}$.
\eit
Let $x = \eplut{\act}$, $y = \eplut{\actb}$, and $u = \utf(\csq)$.  Then
$x, y \in \Eut(\Stt)$, $u \in \ran(\utf)$, 
$x \els_\eudom y$, $\eplut{\act} = x$, $\eplut{\actb} = y$, and 
$\utf(\csq) = u$, 
so (by A6) there
exists 
a partition
$Z_1, \ldots, Z_n$ of $\Stt$, such that $x$ can be expressed as
$x_1 \ep \cdots \ep x_n$ and 
$y$ can be expressed as $y_1 \ep \cdots \ep y_n$, where 
$x_i = \eplutcond{\act}{Z_i}$ and 
$y_i = \eplutcond{\actb}{Z_i}$ for $1 \leq i \leq n$,
and
for all 
$1 \leq k \leq n$, 
\bit
\bl[3.]
\cond{if 
$\plf(Z_k) \et u \ep \EP_{i \neq k} x_i \in \Eut(\Stt)$ 
then}
$\plf(Z_k) \et u \ep \EP_{i \neq k} x_i \els_\eudom y$, 
and
\bl[4.]
\cond{if
$\plf(Z_k) \et u \ep \EP_{i \neq k} y_i \in \Eut(\Stt)$ 
then}
$x \els_\eudom \plf(Z_k) \et u \ep \EP_{i \neq k} y_i$.
\eit
To see that 1 holds,
note that
$\plf(Z_k) \et u \ep \EP_{i \neq k} x_i = \eplut{\lotto{\csq}{Z_{i}}{\act}}$.
\cond{Suppose that $\lotto{\csq}{Z_{i}}{\act} \in \Act$; then 
$\eplut{\lotto{\csq}{Z_{i}}{\act}} \in \Eut(\Stt)$.}
By 3, 
$\plf(Z_k) \et u \ep \EP_{i \neq k} x_i \els_\eudom y$. Thus 
$\lotto{\csq}{Z_{i}}{\act} \gls \actb$ as desired.  The argument that 2 holds
is completely analogous (we use 4 instead of 3 to establish that 2 holds), and
we leave it to the reader. 

Now suppose that $(\ACT,\mr{\gseu}(\dprb))$ \satisfies\ P6. We need to show that 
$\BB{\dprb}$ \satisfies\ A6.  \cond{For this direction we assume that $\dprb$ is \whole.}
Let $x, y \in \Eut(\Stt)$ and $u \in \ran(\utf)$.
Suppose that $x \els_\eudom y$.
Let $\act, \actb \in \Act$ and $\csq \in \Csq$ be such that
 $\eplut{\act} = x$,
$\eplut{\actb} = y$, and $\utf(\csq) = u$. 
We need to show that 
there exists a partition 
$Z_1, \ldots, Z_n$ of $\Stt$,
such that 
 $x$ can be expressed as
$x_1 \ep \cdots \ep x_n$ and 
$y$ can be expressed as $y_1 \ep \cdots \ep y_n$,
where 
$x_i = \eplutcond{\act}{Z_i}$ and 
$y_i = \eplutcond{\actb}{Z_i}$ for $1 \leq i \leq n$,
and
for all 
$1 \leq k \leq n$, 3 and 4 hold.

Since $x \els_\eudom y$, $\act \gls \actb$,
so (by P6) there exists a partition
$Z_1, \ldots, Z_n$ of $\Stt$ such that
 for all $Z_i$, 
1 and 2 hold.
To see that 4 holds, 
note that $\eplut{\lotto{\csq}{Z_{i}}{\actb}} = \plf(Z_k) \et u \ep \EP_{i \neq k} y_i$.
\cond{Suppose that $\eplut{\lotto{\csq}{Z_{i}}{\actb}} \in \Eut(\Stt)$;
then
 $\lotto{\csq}{Z_{i}}{\actb} \in \Act$, since $\dprb$ is \whole.}
By 2, $\act \lsact \lotto{\csq}{Z_{i}}{\actb}$.
Thus
$x \els_\eudom \plf(Z_k) \et u \ep \EP_{i \neq k} y_i$.
The argument that 3 holds is completely analogous (we use 1 instead of 2 to
establish that 3 holds), and we leave that to the reader.

\commentout{
Let $u = \utf(\csq)$; clearly $u \in \ran(\utf)$.
Let $x = \eplut{\act}$ and $y = \eplut{\actb}$.  
Clearly $x, y \in \Eut(\Stt)$.
Since $\act \gls \actb$, 
$x \els_\eudom y$.  
Since $\BB{\dprb}$ \satisfies\ A6, there exists a partition 
$Z_1, \ldots, Z_n$ of $\Stt$,
such that 
$x = x_1 \ep \cdots \ep x_n$ and $y = y_1 \ep \cdots \ep y_n$ for 
$x_i,y_i \in \Eut(Z_i)$, and
\bit
\bl 
$(x \emi x_k)  \ep \plf(Z_k) \et u \els_\eudom y$ and
$x \els_\eudom (y \emi y_k) \ep \plf(Z_k) \et u$.
\eit
Since $\eplut{\lotto{\csq}{Z_{k}}{\act}} = (x \emi x_k)  \ep \plf(Z_k) \et u$ and
$\eplut{\lotto{\csq}{Z_k}{\actb}} = (y \emi y_k) \ep \plf(Z_k) \et u$,
the above implies that
\bit
\bl
$\lotto{\csq}{Z_{i}}{\act} \gls \actb$ 
and 
$\act \gls \lotto{\csq}{Z_{i}}{\actb}$.  
\eit
Thus $(\ACT,\mr{\gseu}(\dprb))$ \satisfies\ P6.
}%

\commentout{
Now suppose that $(\ACT,\mr{\gseu}(\dprb))$ \satisfies\ P6. We need to show that 
$\BB{\dprb}$ \satisfies\ A6.  
Suppose that $x_1, x_2 \in \Eut(\Stt)$ and 
$x_1 \els_\eudom x_2$.  Let $u \in \ran(\utf)$ be given.  We
need to show that 
there exists a partition
$Z_1, \ldots, Z_n$ of $\Stt$ 
all of whose elements are nonempty
such that 
$x = x_1 \ep \cdots \ep x_n$ and $y = y_1 \ep \cdots \ep y_n$ for 
$x_i,y_i \in \Eut(Z_i)$, and
\bit
\bl 
$(x \emi x_k)  \ep \plf(Z_k) \et u \els_\eudom y$ and
$x \els_\eudom (y \emi y_k) \ep \plf(Z_k) \et u$.
\eit
Let $\csq \in \Csq$ be such that $\utf(\csq) = u$ and let 
$\act, \actb \in \Act$ be such that $\eplut{\act} = x$ and $\eplut{\actb} = y$.  
Note that $\act_1 \gls \act_2$; since $(\ACT,\mr{\gseu}(\dprb))$ \satisfies\ 
P6, there exists a partition $Z_1, \ldots, Z_n$ of $\Stt$ 
all of whose elements are nonempty
such that 
for all $Z_i$, 
\bit
\bl
$\lotto{\csq}{Z_{i}}{\act_1} \gls \act_2$ 
and 
$\act_1 \gls \lotto{\csq}{Z_{i}}{\act_2}$.
\eit
Let $x_i = \eplutcond{\act}{Z_i}$ and $y_i = \eplutcond{\actb}{Z_i}$.
Thus,
$x = x_1 \ep \cdots \ep x_n$, $y = y_1 \ep \cdots \ep y_n$, and $x_i,y_i \in \Eut(Z_i)$.
Furthermore, 
$\eplut{\lotto{\csq}{Z_{k}}{\act}} = (x \emi x_k)  \ep \plf(Z_k) \et u$ and
$\eplut{\lotto{\csq}{Z_k}{\actb}} = (y \emi y_k) \ep \plf(Z_k) \et u$.
Thus we have that
\bit
\bl 
$(x \emi x_k)  \ep \plf(Z_k) \et u \els_\eudom y$ and
$x \els_\eudom (y \emi y_k) \ep \plf(Z_k) \et u$.
\eit
So $\dprb$ \satisfies\ A6.
}%
\eit
So far we have 
shown
that $\mr{A}i$ \represents\ $\mr{P}i$
with respect to $\dprbs_i$, for 
$i \in \{\mr{1a}, \mr{1b}, \ldots,  \mr{6}\}$.
Let 
$i_1, \ldots, i_k \in \{\mr{1a}, \mr{1b}, \ldots, \mr{6}\}$. 
Suppose that $\dprb \in \dprbs_{i_1} \cap \cdots \cap \dprbs_{i_k}$
and that $\dprb$
\satisfies\ $\{\mr{A}{i_1}, \ldots, \mr{A}{i_k}\}$.
Since
$\dprb \in \dprbs_{i_j}$ and
$\BB{\dprb}$
\satisfies\ $\mr{A}i_j$, 
it follows that 
$(\ACT,\mr{\gseu}(\dprb))$ \satisfies\ $\mr{P}i_j$.
Thus $(\ACT,\mr{\gseu}(\dprb))$ \satisfies\ $\{\mr{P}{i_1}, \ldots, \mr{P}{i_k}\}$.
Conversely, if 
$\dprb \in \dprbs_{i_1} \cap \cdots \cap \dprbs_{i_k}$ and
$(\ACT,\mr{\gseu}(\dprb))$ \satisfies\ 
$\{\mr{P}{i_1}, \ldots, \mr{P}{i_k}\}$, then 
$\BB{\dprb}$ \satisfies\ $\{\mr{A}{i_1}, \ldots, \mr{A}{i_k}\}$.
Thus
$\{\mr{A}{i_1}, \ldots, \mr{A}{i_k}\}$ \represents\ 
$\{\mr{P}{i_1}, \ldots, \mr{P}{i_k}\}$
with respect to $\dprbs_{i_1} \cap \cdots \cap \dprbs_{i_k}$.
\eprf

\bibliographystyle{chicago}
\bibliography{z,joe,refs,newutil}
\end{document}